\pdfoutput=1

\documentclass[11pt]{article}

\usepackage[preprint]{acl}

\usepackage{times}
\usepackage{latexsym}

\usepackage[T1]{fontenc}

\usepackage[utf8]{inputenc}

\usepackage{microtype}

\usepackage{inconsolata}

\usepackage{graphicx}

\usepackage{hyperref}
\usepackage{url}
\usepackage{booktabs}       
\usepackage{amsfonts}       
\usepackage{nicefrac}       
\usepackage{microtype}      
\usepackage{xcolor}         
\usepackage{wrapfig}

\usepackage{times}
\usepackage{latexsym}
\usepackage{booktabs}
\usepackage{makecell}
\usepackage{multirow}
\usepackage{multicol}
\usepackage{graphicx}
\usepackage{amsmath}
\usepackage{amsfonts}
\usepackage{xcolor}
\usepackage{todonotes}
\usepackage{listings}
\usepackage{subcaption}
\usepackage{wrapfig}

\usepackage{colortbl}

\usepackage{amsmath}
\usepackage{amssymb}
\usepackage{bm}

\usepackage{tcolorbox}
\usepackage[linesnumbered,ruled,vlined]{algorithm2e}

\newcommand{\method}{FourierAttention}
\newcommand{\kernel}{FlashFourierAttention}

\title{Beyond Homogeneous Attention: Memory-Efficient LLMs\\ via Fourier-Approximated KV Cache}

\author{
Xiaoran Liu\textsuperscript{1,2}\thanks{\ \ Equal contribution.}, 
Siyang He\textsuperscript{1}\footnotemark[1],  
Qiqi Wang\textsuperscript{1}\footnotemark[1],  
Ruixiao Li\textsuperscript{1,2}\footnotemark[1],  
Yuerong Song\textsuperscript{1,2}, 
Zhigeng Liu\textsuperscript{1},\\
\textbf{Mianqiu Huang\textsuperscript{1}, 
Linlin Li\textsuperscript{3},
Qun Liu\textsuperscript{3},
Zengfeng Huang\textsuperscript{1,2},
Qipeng Guo\textsuperscript{2,4},
Ziwei He\textsuperscript{2}\thanks{\ \ Corresponding Author. },
Xipeng Qiu\textsuperscript{1,2}\footnotemark[2]}\\
\textsuperscript{1}School of Computer Science, Fudan University, \textsuperscript{2}Shanghai Innovation Institute, \\
\textsuperscript{3}Huawei Noah's Ark Lab, \textsuperscript{4}Shanghai AI Lab\\
\texttt{xrliu24@m.fudan.edu.cn}, 
\texttt{xpqiu@fudan.edu.cn}, \texttt{ziwei.he@sjtu.edu.cn}\\
}

\begin{document}
\maketitle
\begin{abstract}

Large Language Models struggle with memory demands from the growing Key-Value (KV) cache as context lengths increase. Existing compression methods homogenize head dimensions or rely on attention-guided token pruning, often sacrificing accuracy or introducing computational overhead. We propose \textbf{\method}, a training-free framework that exploits the heterogeneous roles of transformer head dimensions: lower dimensions prioritize local context, while upper ones capture long-range dependencies. By projecting the long-context-insensitive dimensions onto orthogonal Fourier bases, {\method} approximates their temporal evolution with fixed-length spectral coefficients. Evaluations on LLaMA models show {\method} achieves the best long-context accuracy on LongBench and Needle-In-A-Haystack (NIAH). Besides, a custom Triton kernel, {\kernel}, is designed to optimize memory via streamlined read-write operations, enabling efficient deployment without performance compromise.

\end{abstract}

\section{Introduction}\label{sec:intro}

\begin{figure*}[!t]
    \centering
    \includegraphics[width=\textwidth]{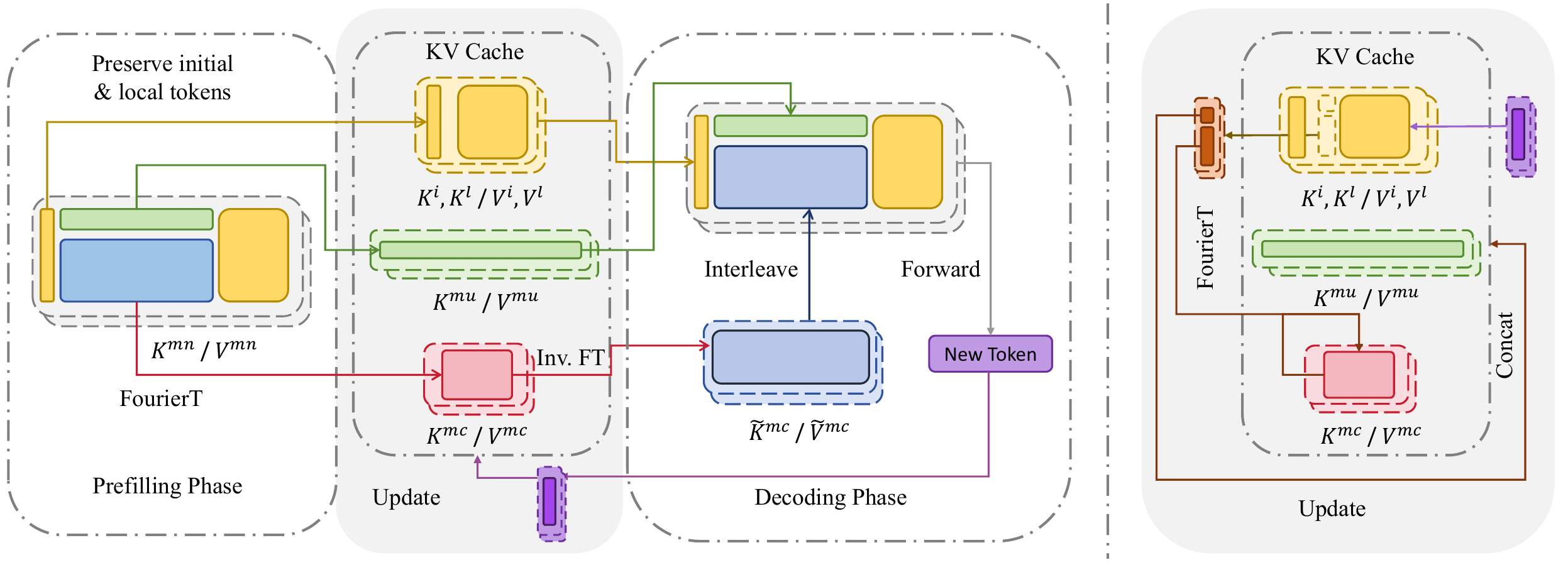}
    \caption{Overview of {\method}.}
    \label{fig:method}
\end{figure*}

Large Language Models (LLMs) have transformed natural language processing with breakthroughs in text generation, comprehension, and reasoning~\citep{gpt4,Sun2024MOSS,OpenAI2024o1,guo2025deepseekr1}. However, their autoregressive decoding relies heavily on a memory-intensive Key-Value (KV) cache, leading to significant memory allocation as context lengths scale~\citep{Vaswani2017attention,fu2024challenges,liu2025thus}. This overhead limits LLM deployment in resource-constrained environments. While approaches like quantization and sparse attention have been explored to reduce memory needs, they often compromise accuracy or add complexity~\citep{liu2024kivi,hooper2024kvquant,yuan2025native}. Developing memory-efficient methods that preserve performance remains critical for broader LLM applicability.


Existing training-free KV cache compression methods, like token eviction strategies~\citep{xiaoefficient,zhang2023h2o,li2024snapkv}, prune sequence subsets but overlook the heterogeneous roles of head dimensions, leaving dimension-aware allocation largely unexplored. Similarly, quantization methods~\citep{liu2024kivi,hooper2024kvquant,duanmu2024skvq} reduce memory by fixed bit-widths, and hidden dimension compression~\citep{chang2024palu,saxena2024eigen} methods apply uniform ratios, both neglect their distinct contribution across dimensions~\citep{liuscaling,pengyarn}. These approaches treat head dimensions as homogeneous, static units rather than dynamically allocating resources based on their importance.


Another critical limitation of existing methods lies in their reliance on attention-guided strategies~\citep{zhang2023h2o,li2024snapkv}. While these approaches enable selective token pruning with minimal accuracy degradation, they impose prohibitive memory and latency overheads due to attention score recalculation. We address this challenge by adapting the HiPPO framework~\citep{gu2020hippo}, a mathematically grounded approach for long-sequence modeling. HiPPO approximates infinite-length sequences as compact finite states by projecting inputs onto finite-order orthogonal basis functions, such as polynomial bases or Fourier bases~\citep{gu2020hippo,he2023fourier}. This retains global critical and contextually vital patterns while filtering out redundant signals. By leveraging HiPPO’s theoretical foundations, we can bypass attention recomputation entirely, achieving both memory efficiency and computational efficiency.


Building on these insights, we introduce \textbf{\method}, a training-free KV cache compression framework using a translated Fourier transform. Departing from prior methods that uniformly process all head dimensions, {\method} identifies localized, context-insensitive dimensions in KV states and approximates their temporal evolution via a fixed set of orthogonal Fourier basis functions. By retaining only the dominant Fourier coefficients ($k \ll L$, where $L$ is the sequence length), our method projects sequences into a compact spectral representation. Unlike polynomial bases that are widely used in HiPPO, which require recurrent state updates, {\method} exploits the shift-invariance and temporal parallelism of Fourier transforms, allowing for efficient computation in a single pass. During decoding, a customized Triton kernel {\kernel} is used to decompose KV cache states during attention calculation, minimizing memory overhead via streamlined read-write operations.



Our contributions can be summarized as follows:
\begin{itemize}
    \item We reveal a bifurcation in Transformer head dimensions: lower dimensions prioritize local context, while upper ones capture long-range dependencies. This inspires us to compress long-context-insensitive dimensions without sacrificing contextual awareness.
    \item We introduce {\method}, which optimizes KV cache by projecting its temporal evolution onto a fixed set of orthogonal Fourier bases. This method efficiently eliminates redundant components while preserving contextual fidelity, achieving a balance between memory and computational efficiency.
    \item We evaluate {\method}'s performance on the LLaMA Series using LongBench and NIAH. Our {\method} achieves the best long-context performance on average while maintaining lower memory consumption.
\end{itemize}

\section{Related Work}\label{sec:related}

KV cache optimization is a crucial technique for enhancing efficiency in attention-based LLMs~\citep{fu2024challenges,liu2025thus}. As context length increases, the KV cache in LLMs grows linearly, creating substantial memory overhead that becomes a bottleneck for long-context applications. Beyond architectural modifications during pretraining~\citep{ainslie2023gqa,liu2024deepseek}, existing training-free optimization approaches mainly involve token eviction or compression. The former discards tokens based on positional or attention patterns, including StreamingLLM~\citep{xiaoefficient}, H2O~\citep{zhang2023h2o}, SnapKV~\citep{li2024snapkv}, and PyramidKV~\citep{cai2024pyramidkv}, while the latter compresses KV cache through quantization or low-rank projection, such as KIVI~\citep{liu2024kivi}, KVQuant~\citep{hooper2024kvquant}, and Palu~\citep{chang2024palu}. However, these methods lack fine-grained consideration of different head dimensions in KV cache, applying uniform optimization across all dimensions. In contrast, our {\method} compresses most dimensions to a fixed length while preserving long-context-sensitive dimensions, effectively reducing KV cache size while maintaining the original long-context capabilities.





\begin{figure*}[!t]
    \begin{subfigure}[b]{0.48\linewidth}
        \centering
        \includegraphics[width=\linewidth]{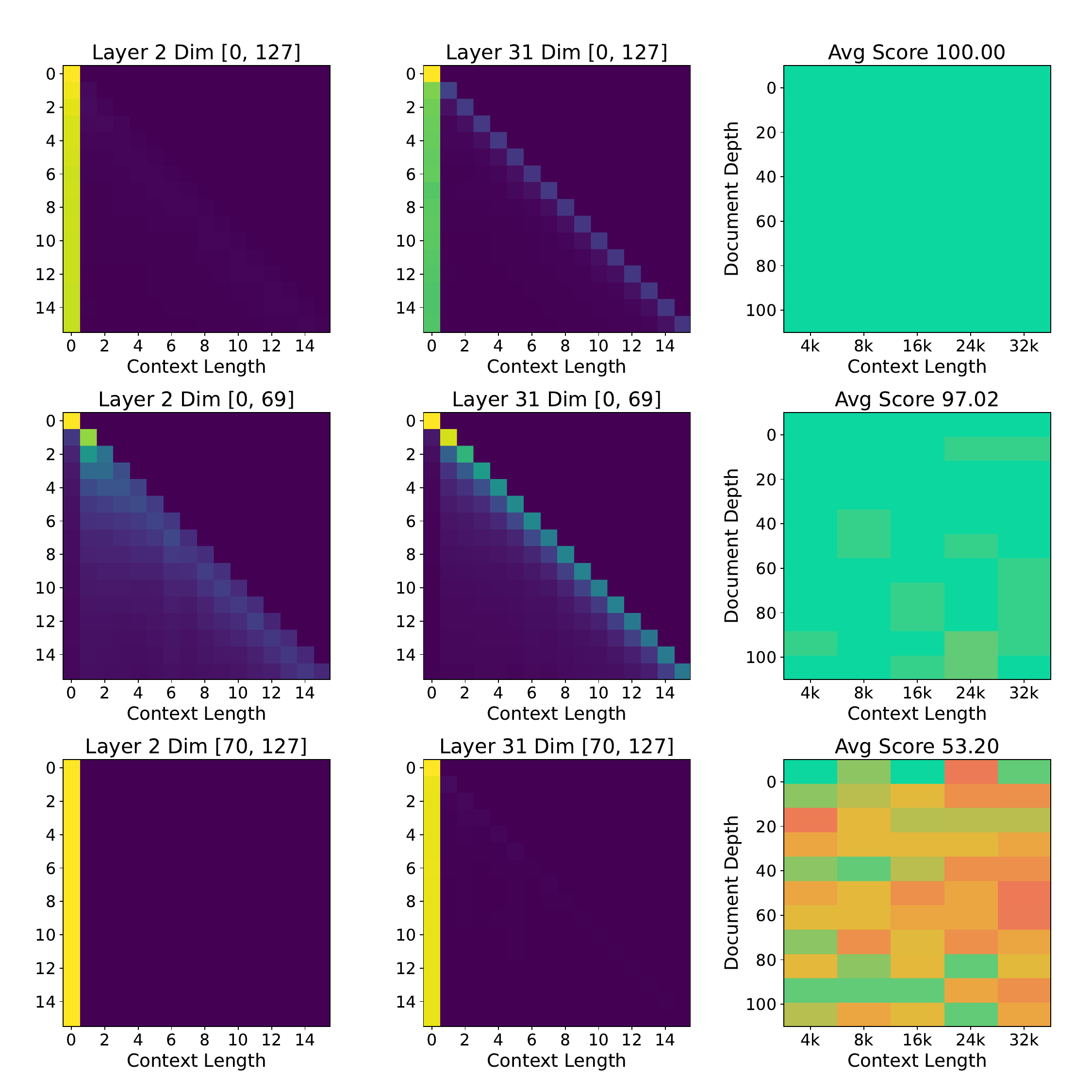}
        \caption{LLaMA3.1-8B}
        \label{llama3_1_8b_heatmap_niah}
    \end{subfigure}
    \hfill
    \begin{subfigure}[b]{0.48\linewidth}
        \centering
        \includegraphics[width=\linewidth]{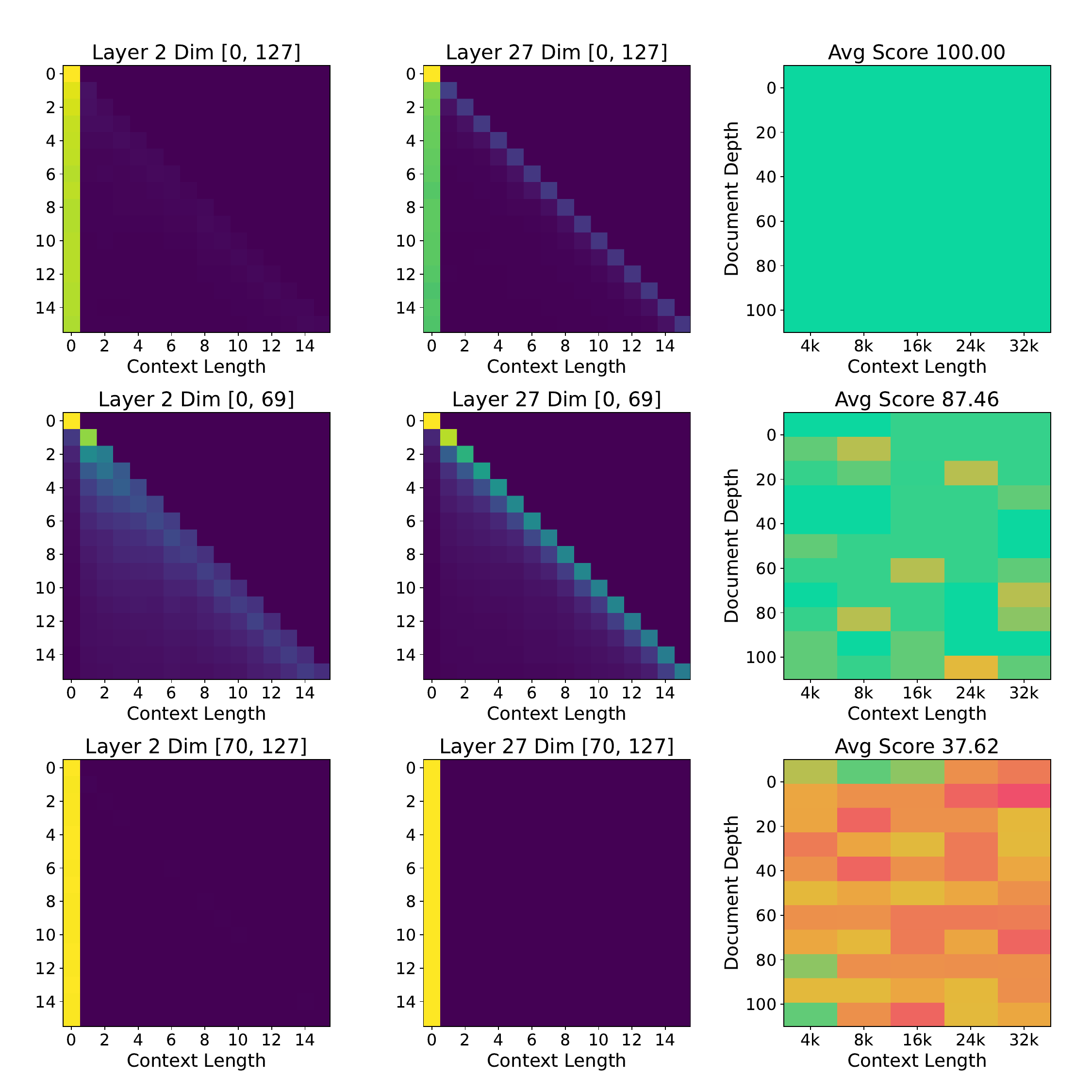}
        \caption{LLaMA3.2-3B}
        \label{llama3_2_3b_heatmap_niah}
    \end{subfigure}
    \caption{Visualization of the average attention score and its components in LLaMA3.1-8B~\citep{dubey2024llama} and LLaMA3.2-3B~\citep{meta2024introducing} over 32 sentences, each with a length of 16. The component of the lower dimensions corresponds to the local branch in StreamingLLM~\citep{xiaoefficient}, while that of the upper dimensions corresponds to the global branch. This reveals the different functions of different dimensions in the attention mechanism. It can be further validated that adding Gaussian noise to the lower dimensions has little effect on NIAH performance, but adding noise to the upper dimensions will harm the performance remarkably.}
    \label{fig:heatmap_niah}
\end{figure*}

\section{Methodology}\label{sec:method}

Due to considerations of different head dimensions of KV cache, we propose \textbf{\method}. In {\method}, most dimensions of the KV cache are compressed to a fixed length through translated Fourier transform, as shown in Figure~\ref{fig:method}. 

\subsection{Head Dimension Specialization}

We analyze the heterogeneous sensitivity of transformer head dimensions to varying context lengths. By visualizing attention scores across 128 dimensions in LLaMA architecture (Figure.\ref{fig:heatmap_niah}), we identify a bifurcation in attention patterns: the first 70 dimensions (0–69) exhibit sharp focus on short-range context, with score distributions concentrated on recent tokens, while the latter 58 dimensions (70–127) maintain a persistent bias toward initial "sink tokens"—positional embeddings that serve as static reference points. This divergence suggests distinct contextual roles encoded within head dimensions, where specialized subsets prioritize local versus global signal retention. 

To further validate this hypothesis, we evaluate the model on a Needle-In-A-Haystack retrieval task across sequences of up to 32,000 tokens. As shown in Figure 1(a), the baseline model achieves perfect retrieval accuracy (100.0). Introducing Gaussian noise to the first 70 dimensions, performance remains robust (97.02), confirming their limited role in long-range dependency resolution. Conversely, perturbing the latter 58 dimensions catastrophically reduces accuracy to 53.20 on average, with failures consistent across all tested depths and context lengths (Figure 1(b) mirrors this trend). This stark contrast empirically demonstrates that upper dimensions in transformer are indispensable for retaining long-range information, while lower dimensions specialize in local context encoding. These findings provide critical insights for optimizing memory-efficient architectures, as strategically prioritizing dimensions specialized in long-range retention enhances contextual awareness within memory limits. For more details on dimension selection, please refer to Section~\ref{sec:select}.

\subsection{Preliminary: HiPPO Framework}\label{sec:hippo}

Inspired by HiPPO~\citep{gu2020hippo}, we compress these less context-sensitive dimensions into fixed-length states to reduce KV cache storage. Under the HiPPO framework, an infinitely long sequence, $f_{1\cdots{L}}$, can be approximated by finite-length states, $\bm{c}\in\mathbb{R}^k$, as the combining coefficients of finite-order basis functions. HiPPO designs different state update equations for various basis functions under different measure functions, such as LegT based on Legendre Polynomial in a translated fixed window size. Among these methods, FourierT measure based on Translated Fourier Transform is most suitable for token-wise parallelism in transformers, because it can be expressed in matrix form and performed independently in different order states. Therefore, we adopt FourierT to compress cache, $\bm{K},\bm{V}\in\mathbb{R}^{L\times{d}}$, which also achieves better downstream performance in Section~\ref{sec:basis}.

\subsection{Online Compression via HiPPO-FourierT}\label{sec:compress}

We set the translated window length in FourierT to the maximum context length, ensuring effective compression within valid input-output ranges. In the prefilling phase, we preserve all dimensions of the initial $L_\text{init}$ and the local $L_\text{local}$ tokens,
\begin{equation}\begin{gathered}
\bm{K}^{i}, \bm{K}^{l} = \bm{K}[:L_\text{init}], \bm{K}[-L_\text{local}:] \\
\bm{V}^{i}, \bm{V}^{l} = \bm{V}[:L_\text{init}], \bm{V}[-L_\text{local}:]
\end{gathered}\end{equation}
and distinguish the dimension indices $\mathcal{D}^{ku}$, $\mathcal{D}^{kc}, \mathcal{D}^{vu}$, $\mathcal{D}^{vc} $ in KV cache for uncompressing and compressing, to enable training-free integration. 
\begin{equation}\begin{aligned}
\bm{K}^{mn}=\bm{K}[L_\text{init}:-L_\text{local}, \mathcal{D}^{kc}], \\
\bm{V}^{mn}=\bm{V}[L_\text{init}:-L_\text{local}, \mathcal{D}^{vc}], \\
\bm{K}^{mu}=\bm{K}[L_\text{init}:-L_\text{local}, \mathcal{D}^{ku}], \\
\bm{V}^{mu}=\bm{V}[L_\text{init}:-L_\text{local}, \mathcal{D}^{vu}].
\end{aligned}\end{equation}
We preserve $\bm{K}^{mu},\bm{V}^{mu}$, compress $\bm{K}^{mn},\bm{V}^{mn}$ to fix-sized $\bm{K}^{mc}\in\mathbb{R}^{2k\times{|\mathcal{D}^{kc}|}},\bm{V}^{mc}\in\mathbb{R}^{2k\times{|\mathcal{D}^{vc}|}}$ and use the original KV for forward propagation.
\begin{equation}\begin{gathered}
\bm{K}^{mc}=\bm{\mathcal{F}}\bm{K}^{mn},\ \ \bm{V}^{mc}=\bm{\mathcal{F}}\bm{V}^{mn},\\[0.5ex]
\bm{O}=\texttt{flash\_attention}(\bm{Q},\bm{K},\bm{V}).
\end{gathered}\end{equation}
When it comes to the compression matrix in FourierT, originally, $\bm{\mathcal{F}}\in\mathbb{C}^{k\times{L}}$, where $\bm{\mathcal{F}}_{nt}=e^{i\frac{2\pi{nt}}{T}}$. However, since caches in mainstream LLMs are real-valued, we convert complex numbers to corresponding 2D vectors, transforming $k$-order complex states into $2k$-order real states. Therefore, the real compression matrix in FourierT is $\bm{\mathcal{F}}\in\mathbb{R}^{2k\times{L}}$ as shown in Equation~\ref{equ:matrix}.
\begin{equation}
    \bm{\mathcal{F}}={\begin{bmatrix}
        1 & 1 & \cdots & 1 \\
        0 & 0 & \cdots & 0 \\
        1 & \cos{\frac{2\pi}{T}} & \cdots & \cos{\frac{2\pi(L-1)}{T}} \\
        0 & \sin{\frac{2\pi}{T}} & \cdots & \sin{\frac{2\pi(L-1)}{T}} \\
        \vdots & \vdots & \ddots & \vdots \\
        1 & \cos{\frac{2\pi{(k-1)}}{T}} & \cdots & \cos{\frac{2\pi{(k-1)(L-1)}}{T}} \\
        0 & \sin{\frac{2\pi{(k-1)}}{T}} & \cdots & \sin{\frac{2\pi{(k-1)(L-1)}}{T}} \\
    \end{bmatrix}}.\label{equ:matrix}
\end{equation}

In the decoding phase, {\method} compresses tokens out of the local range individually,
\begin{equation}\begin{gathered}
\bm{K}^{mc}\leftarrow\bm{K}^{mc}+\bm{f}_{t+1}\bm{K}^{l}[0, \mathcal{D}^{kc}] \\
\bm{V}^{mc}\leftarrow\bm{V}^{mc}+\bm{f}_{t+1}\bm{V}^{l}[0, \mathcal{D}^{vc}] \\
\bm{f}_{t+1} = \begin{bmatrix} 0 & 1 & \cdots & \cos{\frac{2\pi(k-1)t}{T}} & \sin{\frac{2\pi(k-1)t}{T}}\end{bmatrix}^\top
\end{gathered}\end{equation}
and reconstruct intermediate cache $\tilde{\bm{K}}^{m},\tilde{\bm{V}}^{m}$ via inverse Fourier transform in attention computation with the current query vector $\bm{q}_{t+1}$. 
\begin{equation}\begin{gathered}
\tilde{\bm{K}}^{m}[\mathcal{D}^{ku}]\leftarrow\bm{K}^{mu},\ \tilde{\bm{K}}^{m}[\mathcal{D}^{kc}]\leftarrow\frac{1}{k}\bm{\mathcal{F}}^T\bm{K}^{mc} \\
\tilde{\bm{K}}=\texttt{cat}(\bm{K}^{i},\tilde{\bm{K}}^{m},\bm{K}^{l}) \\
\tilde{\bm{V}}^{m}[\mathcal{D}^{vu}]\leftarrow\bm{V}^{mu},\ \tilde{\bm{V}}^{m}[\mathcal{D}^{vc}]\leftarrow\frac{1}{k}\bm{\mathcal{F}}^T\bm{K}^{mc} \\
\tilde{\bm{V}}=\texttt{cat}(\bm{V}^{i},\tilde{\bm{V}}^{m},\bm{V}^{l}) \\
\bm{o}_{t+1}=\texttt{flash\_attention}(\bm{q}_{t+1},\tilde{\bm{K}},\tilde{\bm{V}}).
\end{gathered}\end{equation}
To eliminate intermediate read-write cost in decompression, we try to implement a custom kernel, {\kernel}, using Triton~\citep{tillet2019triton}, integrating the decompression into standard FlashAttention2~\citep{daoflashattention} and FlashDecoding~\citep{dao2023flash}. {\kernel} loads compressed intermediate states once and decompresses at corresponding sequence positions during iterative KV cache loading. {\kernel} is still in progress to achieve better computational efficiency compared with standard attention.

\subsection{Fine-Grained Compression Schema}\label{sec:select}

\begin{figure}[!t]
    \begin{subfigure}[b]{0.49\linewidth}
        \centering
        \includegraphics[width=\linewidth]{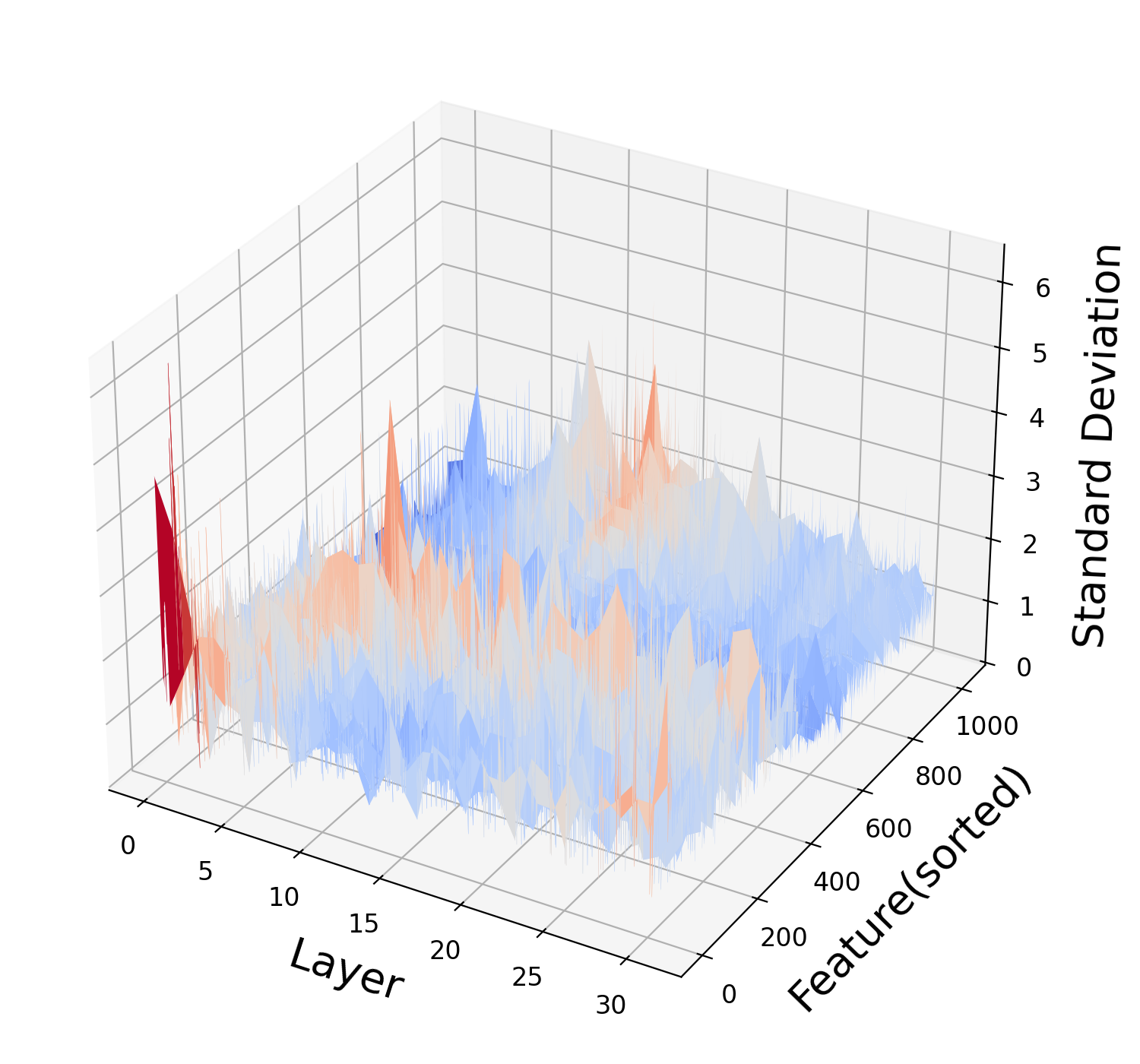}
        \caption{K cahce in LLaMA3.1-8B}
        \label{std_k_llama3_1_8b}
    \end{subfigure}
    \hfill
    \begin{subfigure}[b]{0.49\linewidth}
        \centering
        \includegraphics[width=\linewidth]{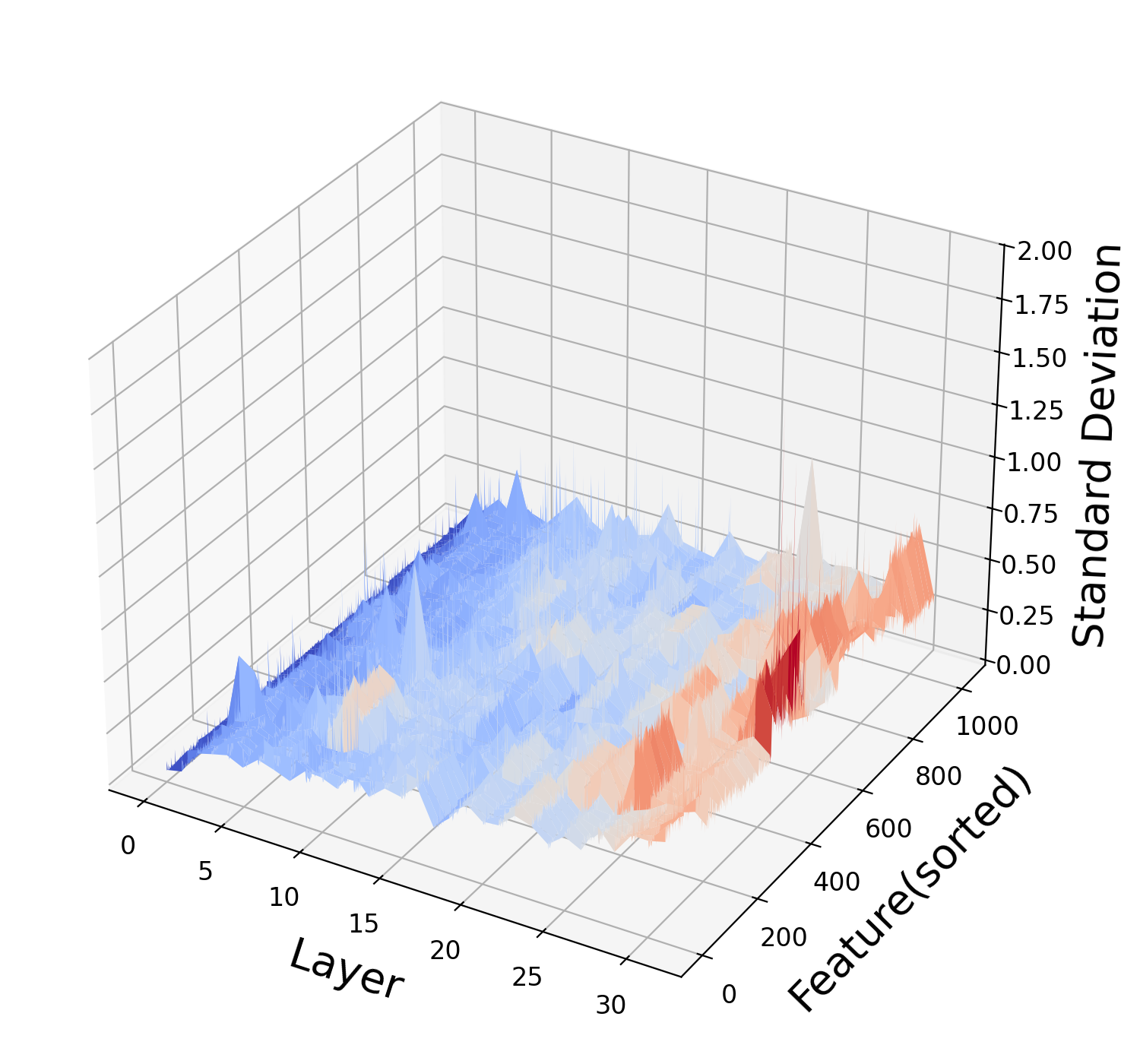}
        \caption{V cahce in LLaMA3.1-8B}
        \label{std_v_llama3_1_8b}
    \end{subfigure}
    \begin{subfigure}[b]{0.49\linewidth}
        \centering
        \includegraphics[width=\linewidth]{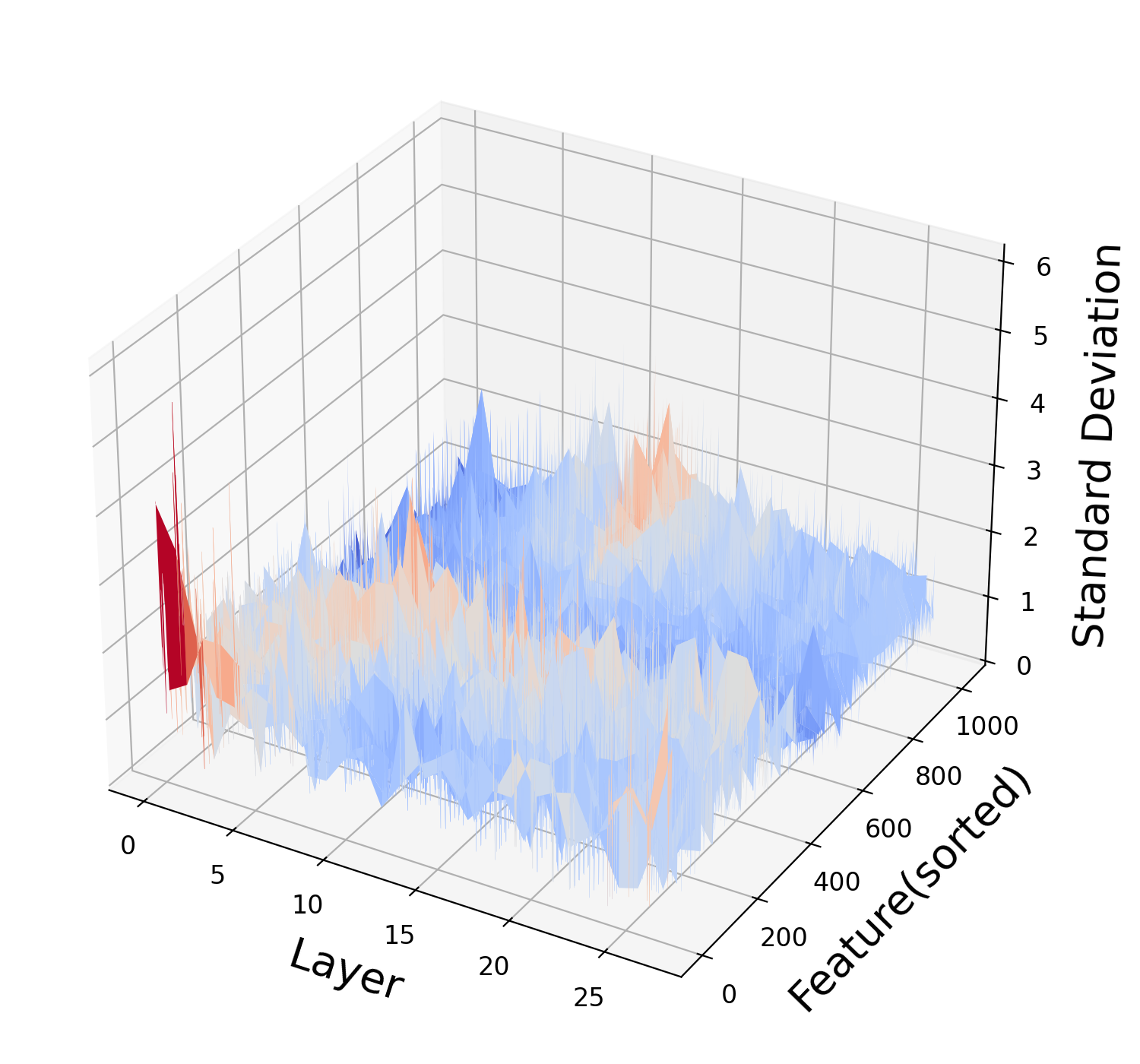}
        \caption{K cahce in LLaMA3.2-3B}
        \label{std_k_llama3_2_3b}
    \end{subfigure}
    \hfill
    \begin{subfigure}[b]{0.49\linewidth}
        \centering
        \includegraphics[width=\linewidth]{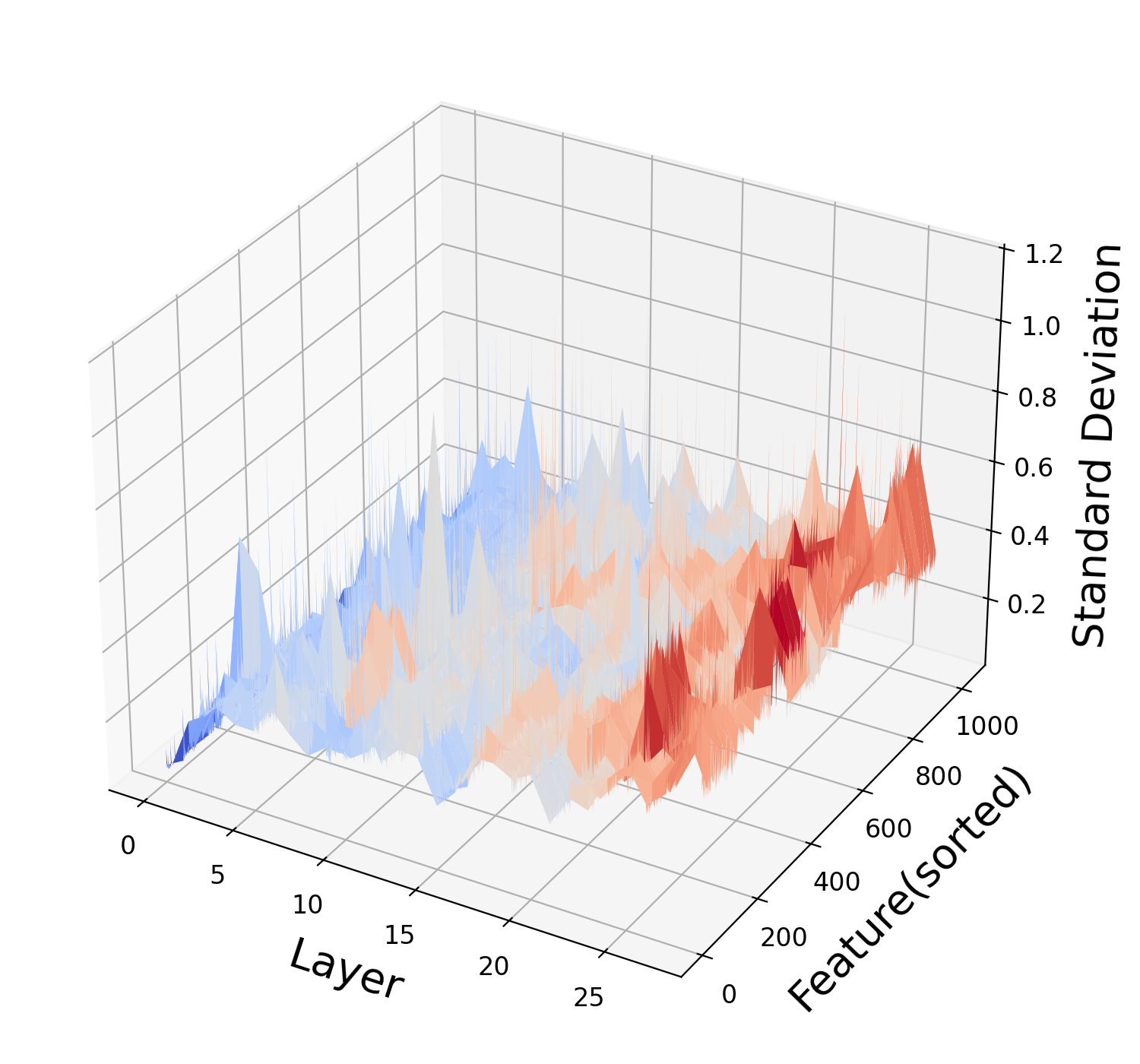}
        \caption{V cahce in LLaMA3.2-3B}
        \label{std_v_llama3_2_3b}
    \end{subfigure}
    \caption {Visualization of standard deviation of KV cache in different layers in LLaMA3.1-8B~\citep{dubey2024llama} and LLaMA3.2-3B~\citep{meta2024introducing}. The feature dimensions are sorted based on the indices in each head.\label{fig:kv_feature}}
\end{figure}

In {\method}, a crucial point lies in how to select the dimension to be compressed. To address this, we directly compress and decompress all KV caches, prioritizing dimensions with smaller mean-squared error in reconstruction to a fixed length. 
Based on further observations of the KV cache, we adopt a fine-grained compression schema, where more dimensions of the V cache and lower-layer caches are compressed to a fixed length. We analyze the standard deviation of KV cache dimensions along the temporal direction across different layers and find that for both LLaMA3.1-8B and LLaMA3.2-3B, as shown in Figure~\ref{fig:kv_feature}. The standard deviation of the K cache is consistently higher than that of the V cache, and the standard deviation in upper layers exceeds that of lower layers. Consequently, we compress more dimensions of the smoother V cache and lower-layer caches to a fixed length, while retaining more K cache and upper-layer caches to extend with sequence length. Thus, {\method} exhibits an asymmetric, inverted-pyramid compression pattern.

Interestingly, this differs from most KV cache compression approaches. Works like \citet{cai2024pyramidkv} and \citet{xing2024pyramiddrop} suggest preserving more KV caches in lower layers, as attention becomes sparser in upper layers. However, in {\method}, the optimization criterion is whether the dimension can be well reconstructed. Since caches in upper layers exhibit more oscillatory features due to more deterministic predictions, we retain more dimensions to maintain output stability.

\section{Experiment}\label{sec:experiment}

\begin{table*}[!tb]
\tabcolsep=0.15cm
\centering\small
\begin{tabular}{l ccc ccc ccc ccc}  %
\toprule
& \multicolumn{3}{c}{\textbf{Single-Doc QA}} & \multicolumn{3}{c}{\textbf{Multi-Doc QA}} & \multicolumn{3}{c}{\textbf{Summarization}} & \multicolumn{3}{c}{\textbf{Few-shot Learning}} \\
\cmidrule(lr){2-4} \cmidrule(lr){5-7} \cmidrule(lr){8-10} \cmidrule(lr){11-13}
& NQA & Qsp & MulF & HQA & 2WQA & MSQ & QRpt & QSum & MulN & TREC & TrQA & SSum \\
\midrule
\textcolor{gray}{\textbf{\textit{LLaMA3.1-8B}}} & \textcolor{gray}{13.22} & \textcolor{gray}{20.24} & \textcolor{gray}{32.81} & \textcolor{gray}{11.97} & \textcolor{gray}{13.60} & \textcolor{gray}{8.72} & \textcolor{gray}{29.70} & \textcolor{gray}{25.09} & \textcolor{gray}{0.90} & \textcolor{gray}{73.50} & \textcolor{gray}{90.97} & \textcolor{gray}{47.26} \\ 
+ StreamingLLM & 7.87 & 13.86 & 15.59 & 7.79 & 10.08 & 4.49 & 19.93 & 21.52 & 9.88 & 61.50 & 84.66 & 43.48 \\ 
+ SnapKV & 12.67 & 19.84 & 32.48 & 12.00 & 13.76 & 8.59 & 29.19 & 24.90 & 12.56 & 73.00 & 90.97 & 46.53 \\
+ Palu & 4.50 & 18.03 & 21.58 & 9.47 & 11.32 & 5.23 & 17.25 & 6.90 & 8.98 & 68.50 & 83.38 & 32.29 \\
+ KIVI & 12.73 & 20.94 & 32.79 & 11.51 & 13.93 & 8.77 & 30.33 & 25.18 & 2.41 & 73.50 & 10.83 & 45.98 \\ 
+ Ours & 14.94 & 18.47 & 30.47 & 13.46 & 14.29 & 9.64 & 22.34 & 22.69 & 5.00 & 72.50 & 89.18 & 43.80 \\
\midrule
\textcolor{gray}{\textbf{\textit{LLaMA3.2-3B}}} & \textcolor{gray}{10.27} & \textcolor{gray}{21.69} & \textcolor{gray}{35.52} & \textcolor{gray}{9.58} & \textcolor{gray}{12.78} & \textcolor{gray}{6.75} & \textcolor{gray}{30.15} & \textcolor{gray}{23.77} & \textcolor{gray}{28.24} & \textcolor{gray}{70.00} & \textcolor{gray}{87.24} & \textcolor{gray}{38.19} \\ 
+ StreamingLLM & 9.14 & 17.59 & 21.57 & 7.06 & 9.78 & 3.99 & 19.02 & 21.32 & 23.23 & 53.00 & 84.33 & 39.76 \\ 
+ SnapKV & 8.92 & 21.04 & 34.97 & 9.50 & 12.77 & 6.62 & 29.45 & 23.36 & 27.77 & 69.50 & 86.39 & 38.33 \\ 
+ Palu & 1.98 & 19.17 & 20.37 & 5.84 & 10.28 & 2.65 & 13.37 & 4.10 & 13.97 & 57.00 & 47.39 & 21.48 \\ 
+ KIVI & 10.21 & 22.17 & 35.08 & 9.68 & 12.29 & 6.94 & 30.84 & 23.52 & 23.21 & 70.00 & 64.09 & 43.46 \\ 
+ Ours & 8.91 & 21.27 & 31.20 & 11.99 & 17.77 & 8.26 & 23.41 & 22.16 & 23.76 & 69.00 & 86.18 & 37.12 \\ 
\bottomrule
  \end{tabular}
  \caption{\label{longbench1} Results of LLaMA Series~\citep{dubey2024llama,meta2024llama} on LongBench~\citep{bai2023longbench}. Our {\method} achieves a superiority over StreamingLLM~\citep{xiaoefficient}, Palu~\citep{chang2024palu}, and KIVI~\citep{liu2024kivi} and shows closest performance with LLMs with SnapKV~\citep{li2024snapkv}.}
\end{table*}

\begin{table}[!thb]
\tabcolsep=0.15cm
\centering\small
\begin{tabular}{l cc cc c}
\toprule
 & \multicolumn{2}{c}{\textbf{Synthetic}} & \multicolumn{2}{c}{\textbf{Code}} & \multirow{2}{*}{\textbf{Avg.}} \\
\cmidrule(lr){2-3} \cmidrule(lr){4-5}
 & PsgC & PsgR & LCC & Re-P &  \\
\midrule
\textcolor{gray}{\textbf{\textit{LLaMA3.1-8B}}} & \textcolor{gray}{0.75} & \textcolor{gray}{26.75} & \textcolor{gray}{72.00} & \textcolor{gray}{69.27} & \textcolor{gray}{39.49} \\
+ StreamingLLM & 1.25 & 6.18 & 58.36 & 56.35 & 31.52 \\
+ SnapKV & 0.75 & 26.75 & 59.97 & 59.74 & \textbf{37.07} \\
+ Palu & 0.60 & 14.58 & 56.68 & 54.57 & 30.68 \\
+ KIVI & 0.75 & 30.00 & 29.43 & 19.86 & 23.18 \\
+ Ours & 2.25 & 16.07 & 67.32 & 63.02 & \underline{36.98} \\
\midrule
\textcolor{gray}{\textbf{\textit{LLaMA3.2-3B}}} & \textcolor{gray}{0.00} & \textcolor{gray}{7.00} & \textcolor{gray}{70.01} & \textcolor{gray}{66.38} & \textcolor{gray}{38.04} \\
+ StreamingLLM & 1.37 & 6.47 & 55.47 & 53.53 & 31.19 \\
+ SnapKV & 0.00 & 6.75 & 58.20 & 56.36 & \underline{34.83} \\
+ Palu & 1.45 & 3.08 & 55.10 & 49.54 & 25.53 \\
+ KIVI & 0.00 & 6.00 & 40.01 & 37.87 & 28.98 \\
+ Ours & 1.62 & 8.32 & 67.29 & 59.91 & \textbf{36.33} \\
\bottomrule
\end{tabular}
  \caption{\label{longbench2} Continuous table of Table \ref{longbench1}.}
\end{table}

\subsection{Setup}\label{sec:setup}

We conduct experiments on LLaMA3.1-8B~\citep{dubey2024llama} and LLaMA3.2-3B~\citep{meta2024introducing}. For all models, we set the length of initial tokens $L_\text{init}$ to 4, the length of local tokens $L_\text{local}$ to 1024, and the number of states $k=512$. We evaluate the reconstruction loss using the prompt portion of the 32k Needle-In-A-Haystack benchmark in OpenCompass~\citep{2023opencompass}. As mentioned earlier, we employ an asymmetric inverted pyramid compression strategy: for the first 4 layers, we compress 90\% of K dimensions and 95\% of V dimensions; for the last 8 layers, 50\% of K and 70\% of V; and for the remaining layers, 80\% of both K and V. Overall, 76\% KV caches are compressed to a fixed length. All experiments are performed on an NVIDIA H100 GPU with FP16 precision and accelerated with FlashAttention2~\citep{daoflashattention}.

\subsection{Long-Context Evaluation}

We evaluate our method against other KV cache optimization approaches with two long-context benchmarks in OpenCompass~\citep{2023opencompass},  LongBench~\citep{bai2023longbench} and Needle-In-A-Haystack (NIAH)~\citep{niah,li2024needlebench}, with a truncation context length of 32K. We compare with StreamingLLM~\citep{xiaoefficient}, SnapKV~\citep{li2024snapkv}, Palu~\citep{chang2024palu}, and KIVI~\citep{liu2024kivi}, covering both token eviction and feature compression. For fair comparison, we retain 4 initial tokens and 1024 local tokens in StreamingLLM, additionally keep 1024 recalled middle tokens, matching our compressed dimension count, in SnapKV, compress KV feature dimensions to 70\% in Palu, and apply 4-bit quantization, 75\% compression, in KIVI.

\begin{figure*}[!tb]
    \begin{subfigure}[b]{0.32\linewidth}
        \centering
        \includegraphics[width=\linewidth]{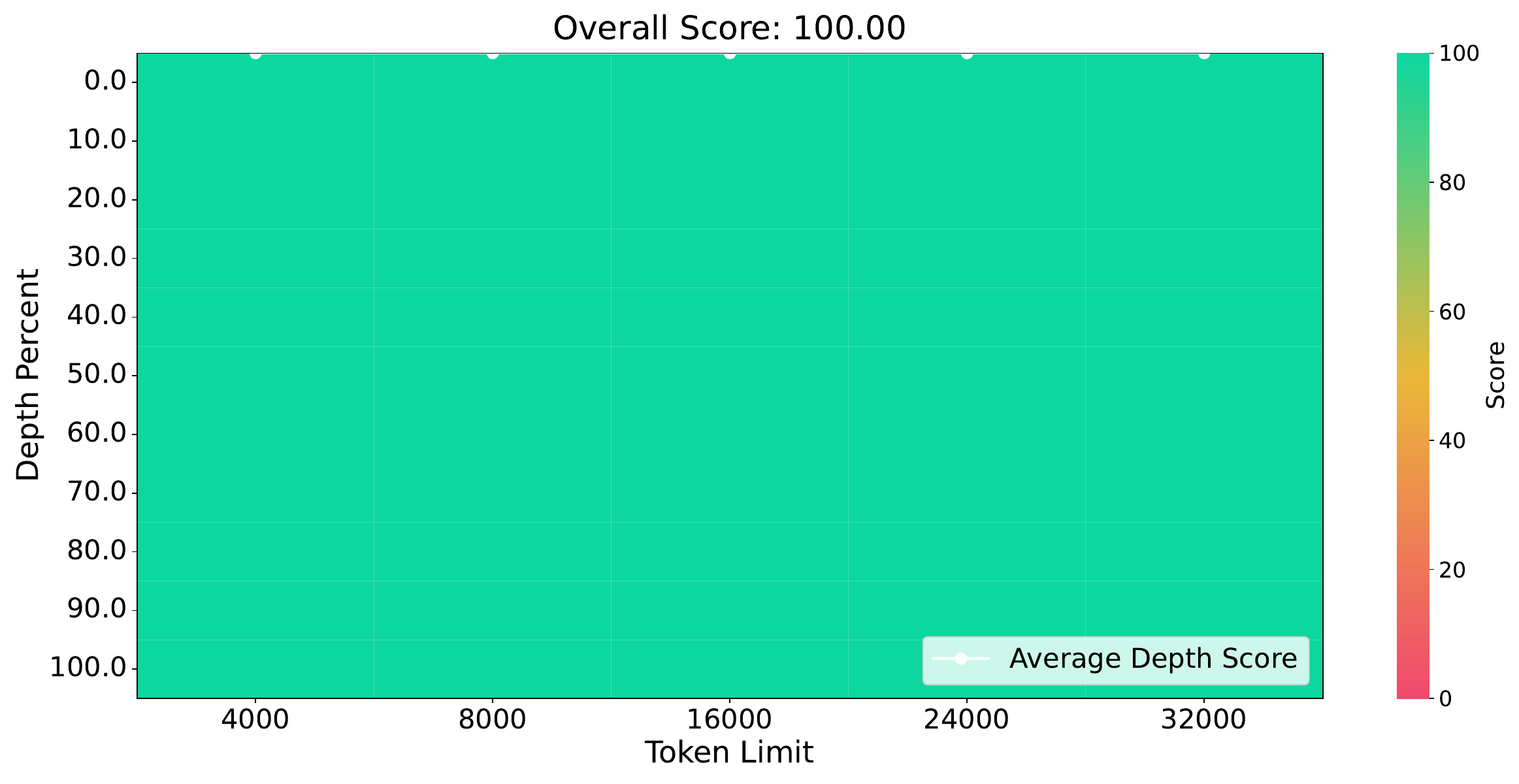}
        \caption{Pretrained}
        \label{niah_llama3_1_8b_pretrained}
    \end{subfigure}
    \hfill
    \begin{subfigure}[b]{0.32\linewidth}
        \centering
        \includegraphics[width=\linewidth]{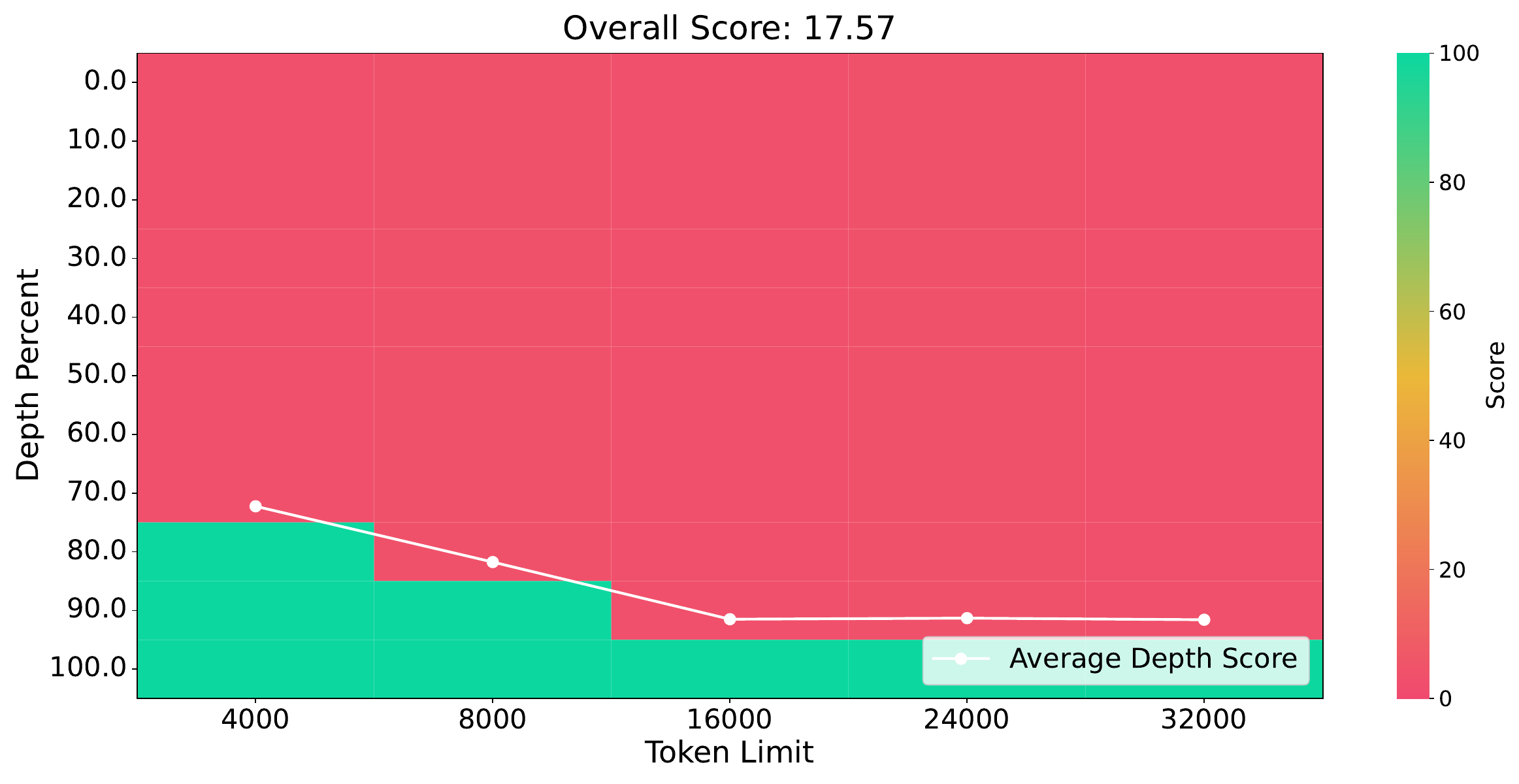}
        \caption{StreamingLLM}
        \label{niah_llama3_1_8b_streamingllm}
    \end{subfigure}
    \hfill
    \begin{subfigure}[b]{0.32\linewidth}
        \centering
        \includegraphics[width=\linewidth]{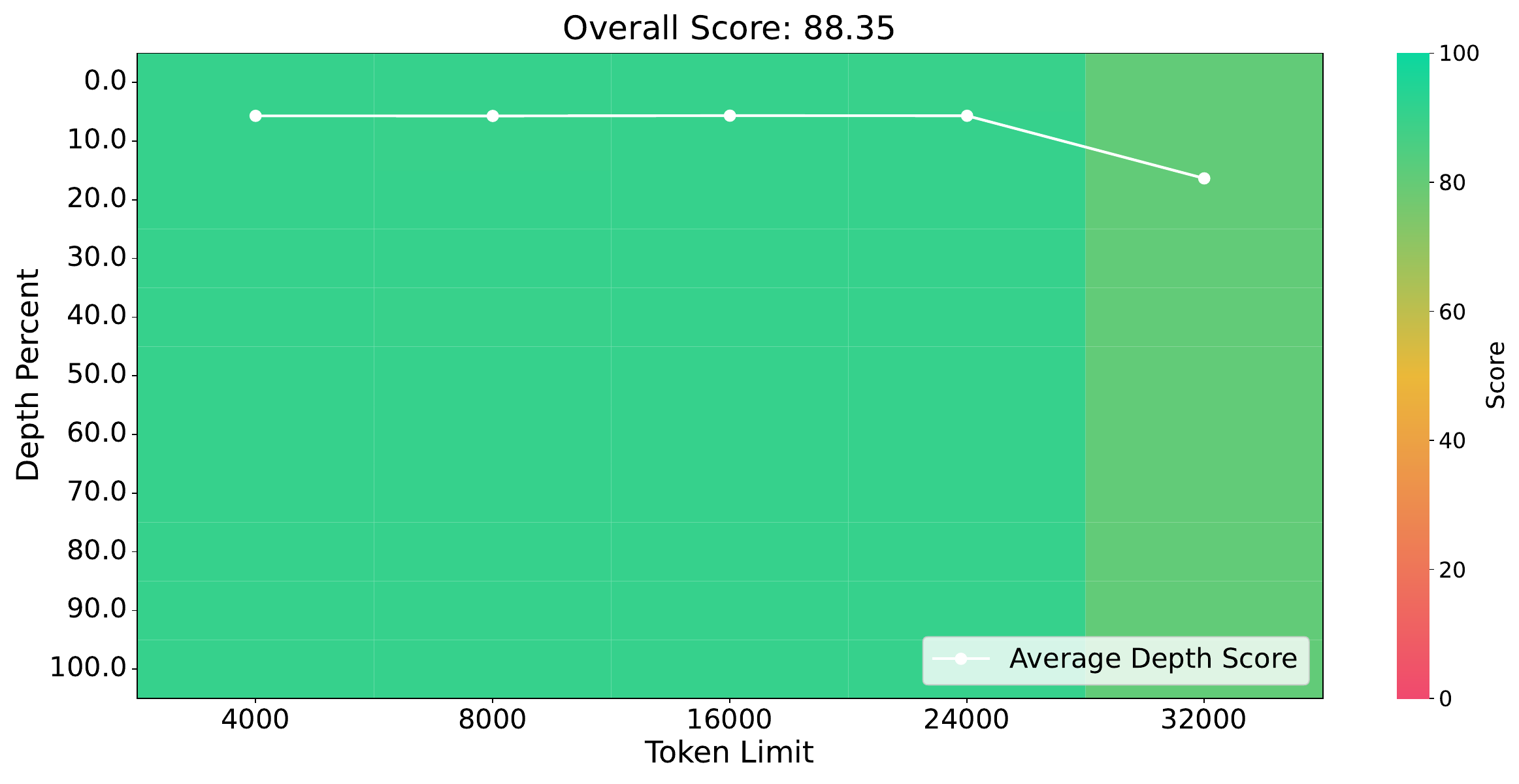}
        \caption{SnapKV}
        \label{niah_llama3_1_8b_snapkv}
    \end{subfigure}
    \vskip\baselineskip
    \begin{subfigure}[b]{0.32\linewidth}
        \centering
        \includegraphics[width=\linewidth]{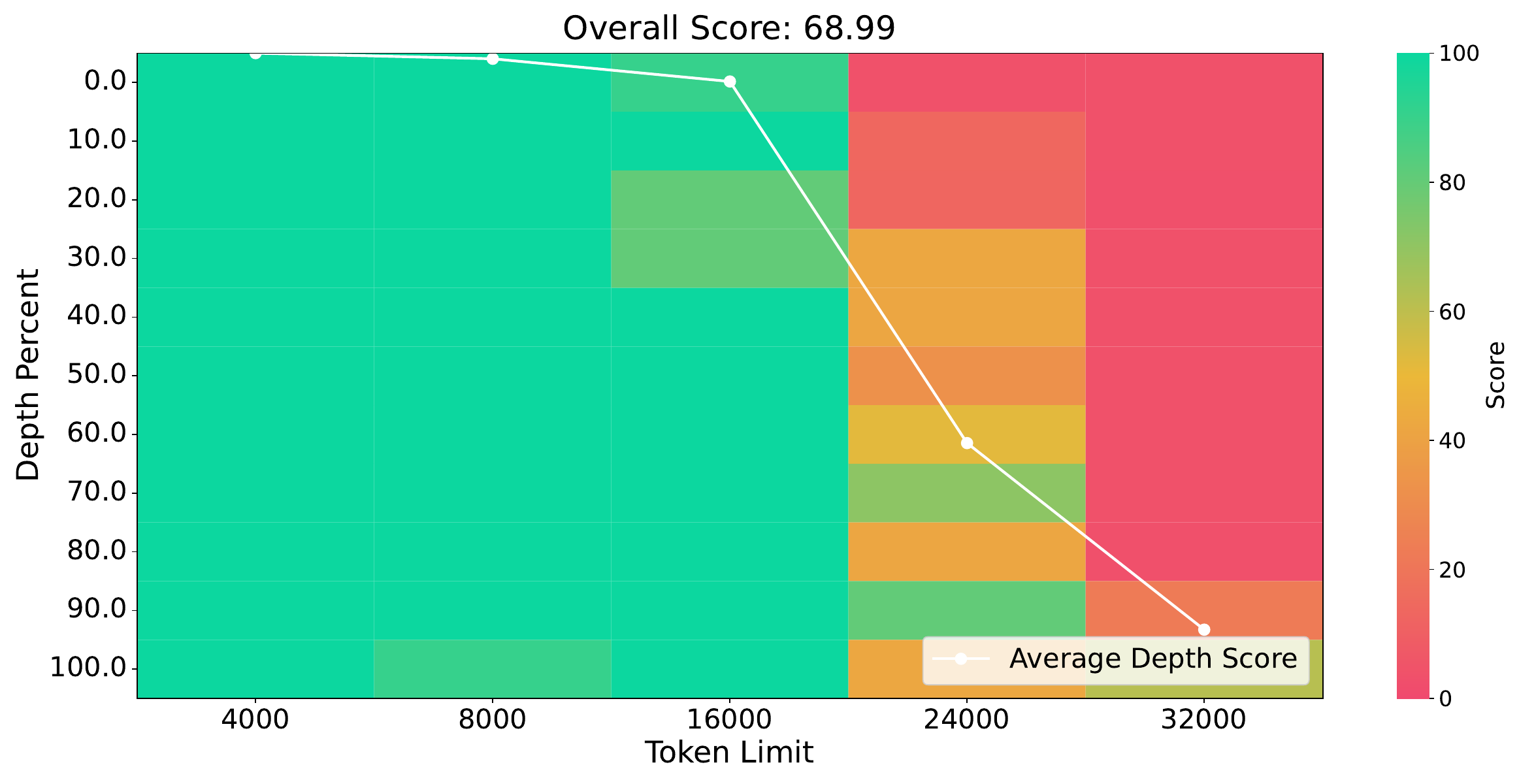}
        \caption{Palu}
        \label{niah_llama3_1_8b_palu}
    \end{subfigure}
    \hfill
    \begin{subfigure}[b]{0.32\linewidth}
        \centering
        \includegraphics[width=\linewidth]{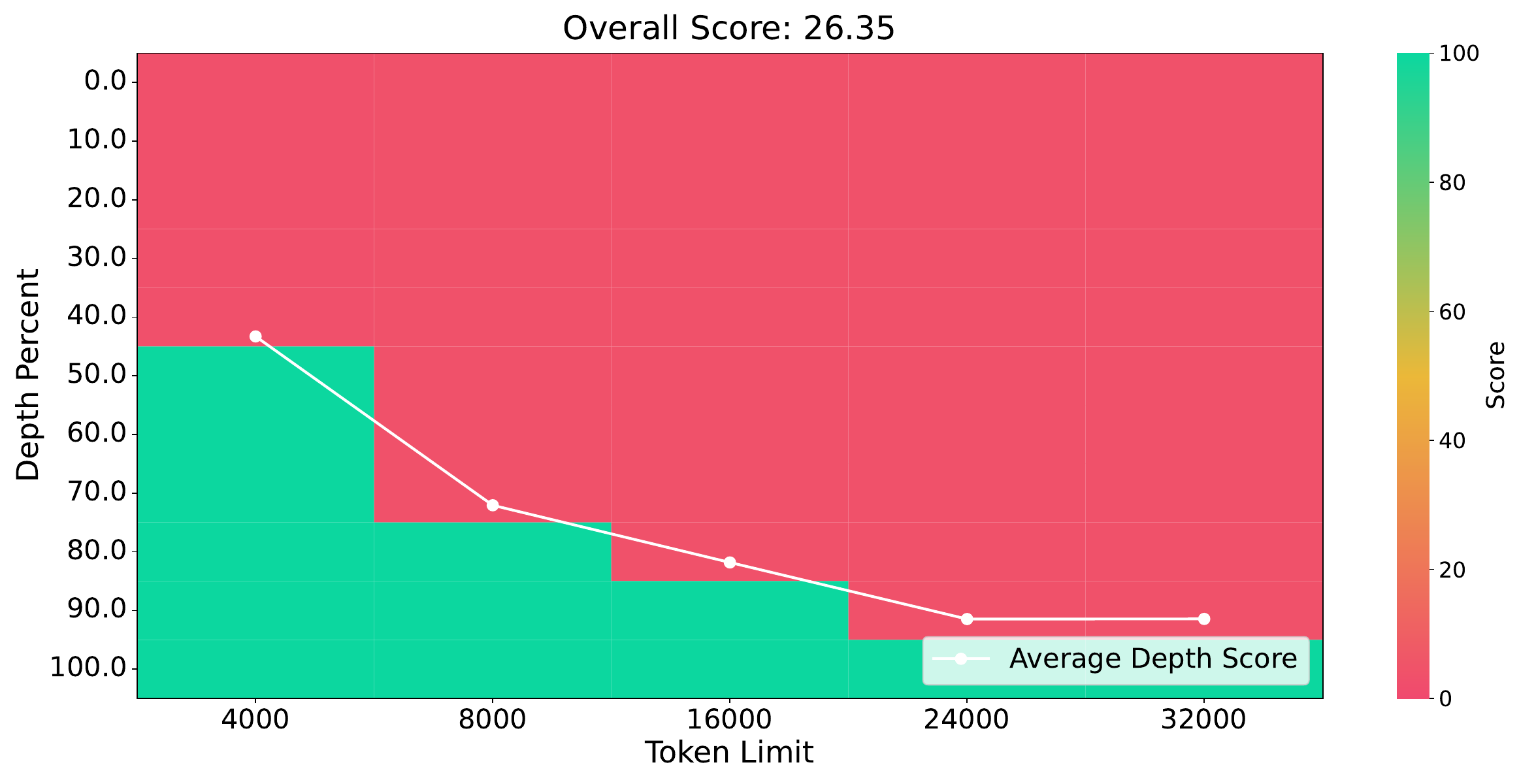}
        \caption{KIVI}
        \label{niah_llama3_1_8b_kivi}
    \end{subfigure}
    \hfill
    \begin{subfigure}[b]{0.32\linewidth}
        \centering
        \includegraphics[width=\linewidth]{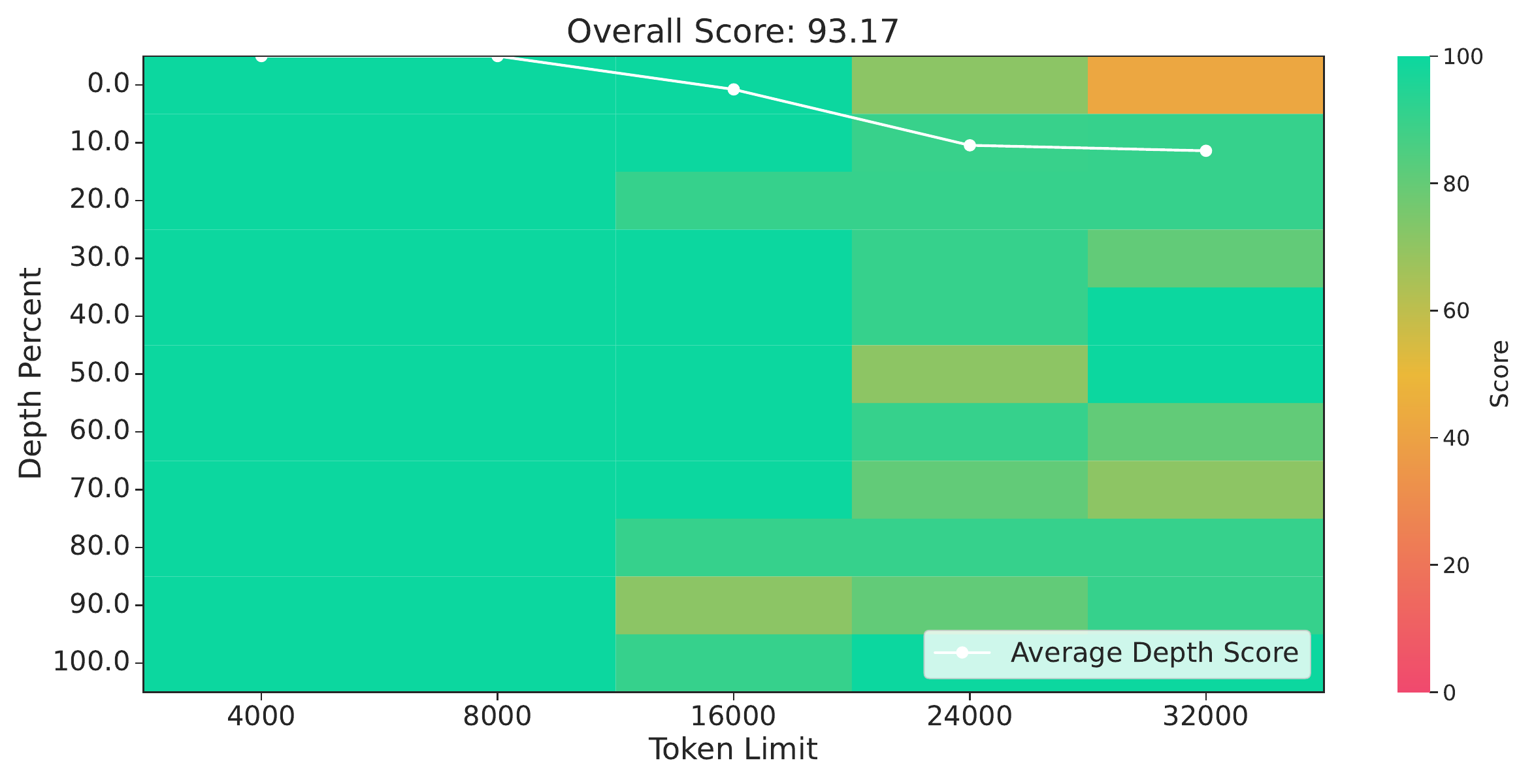}
        \caption{{\method} (ours)}
        \label{niah_llama3_1_8b_ours}
    \end{subfigure}
    \caption{Results of LLaMA3.1-8B~\citep{dubey2024llama} on Needle-In-A-Haystack~\citep{niah}. {\method} achieves a highest average score over StreamingLLM~\citep{xiaoefficient}, SnapKV~\citep{li2024snapkv}, Palu~\citep{chang2024palu}, and KIVI~\citep{liu2024kivi} and shows closest performance with LLMs with full attention. \label{niah_llama3_1_8b}}
\end{figure*}

\begin{figure*}[!tb]
    \begin{subfigure}[b]{0.32\linewidth}
        \centering
        \includegraphics[width=\linewidth]{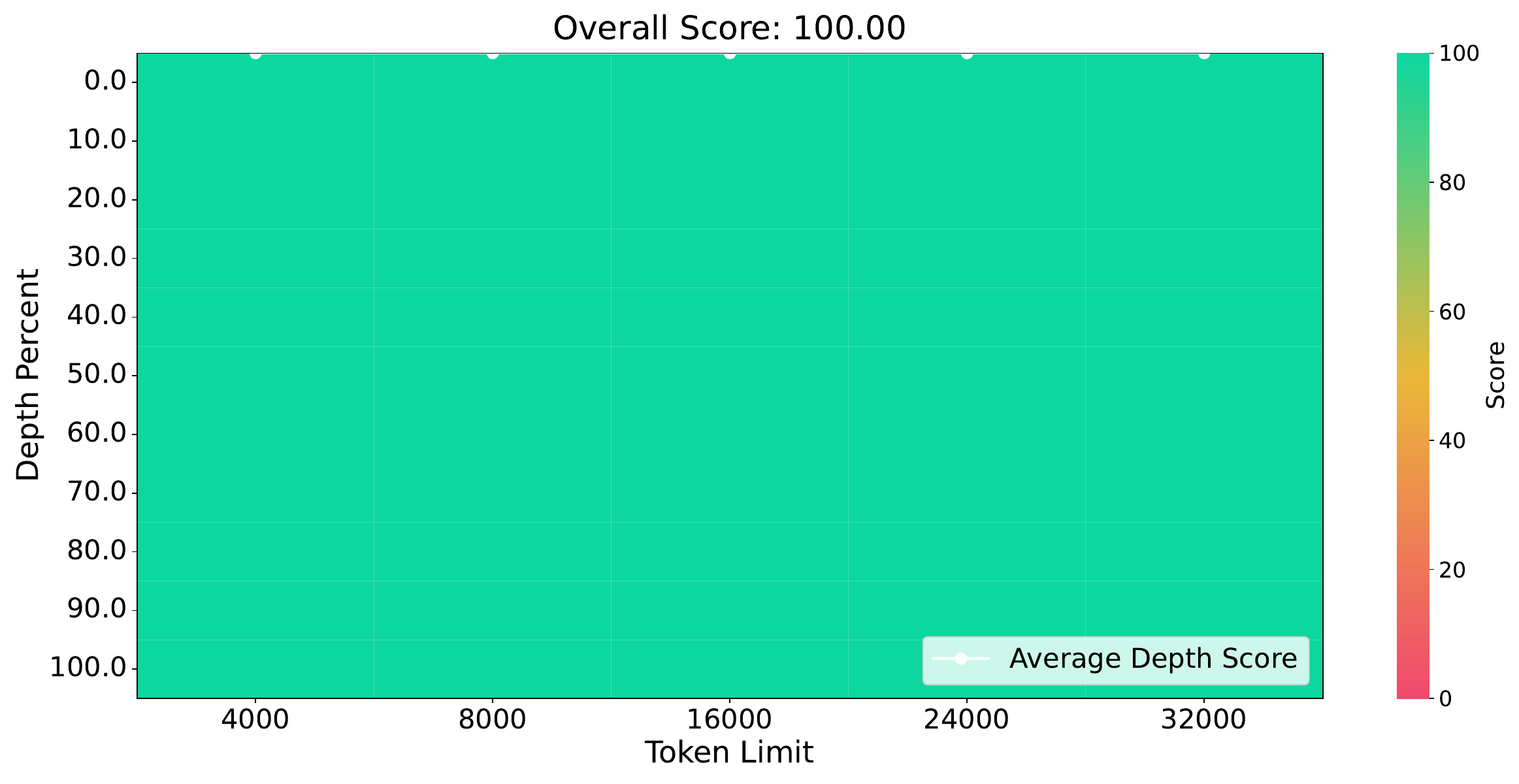}
        \caption{Pretrained}
        \label{niah_llama3_2_3b_pretrained}
    \end{subfigure}
    \hfill
    \begin{subfigure}[b]{0.32\linewidth}
        \centering
        \includegraphics[width=\linewidth]{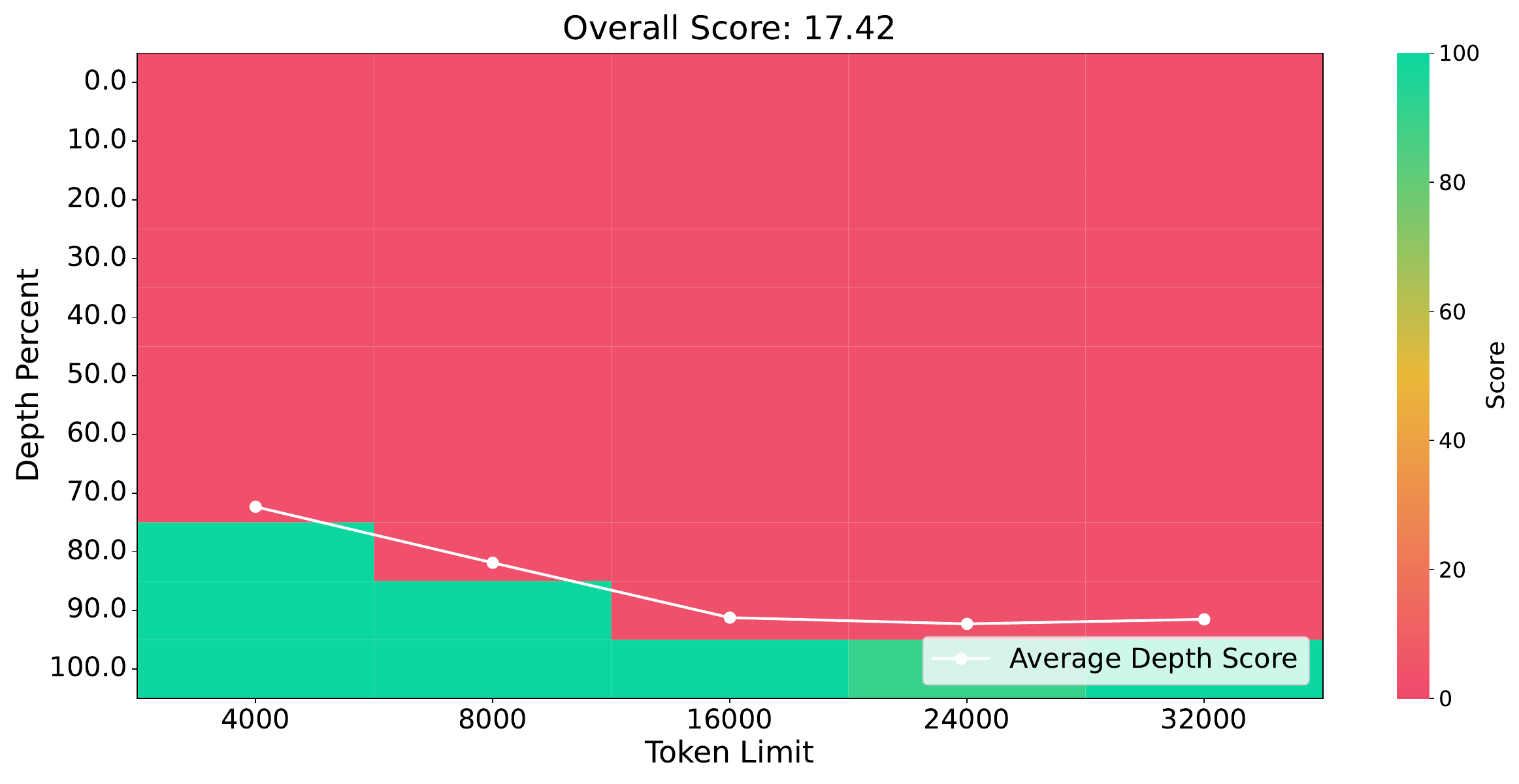}
        \caption{StreamingLLM}
        \label{niah_llama3_2_3b_streamingllm}
    \end{subfigure}
    \hfill
    \begin{subfigure}[b]{0.32\linewidth}
        \centering
        \includegraphics[width=\linewidth]{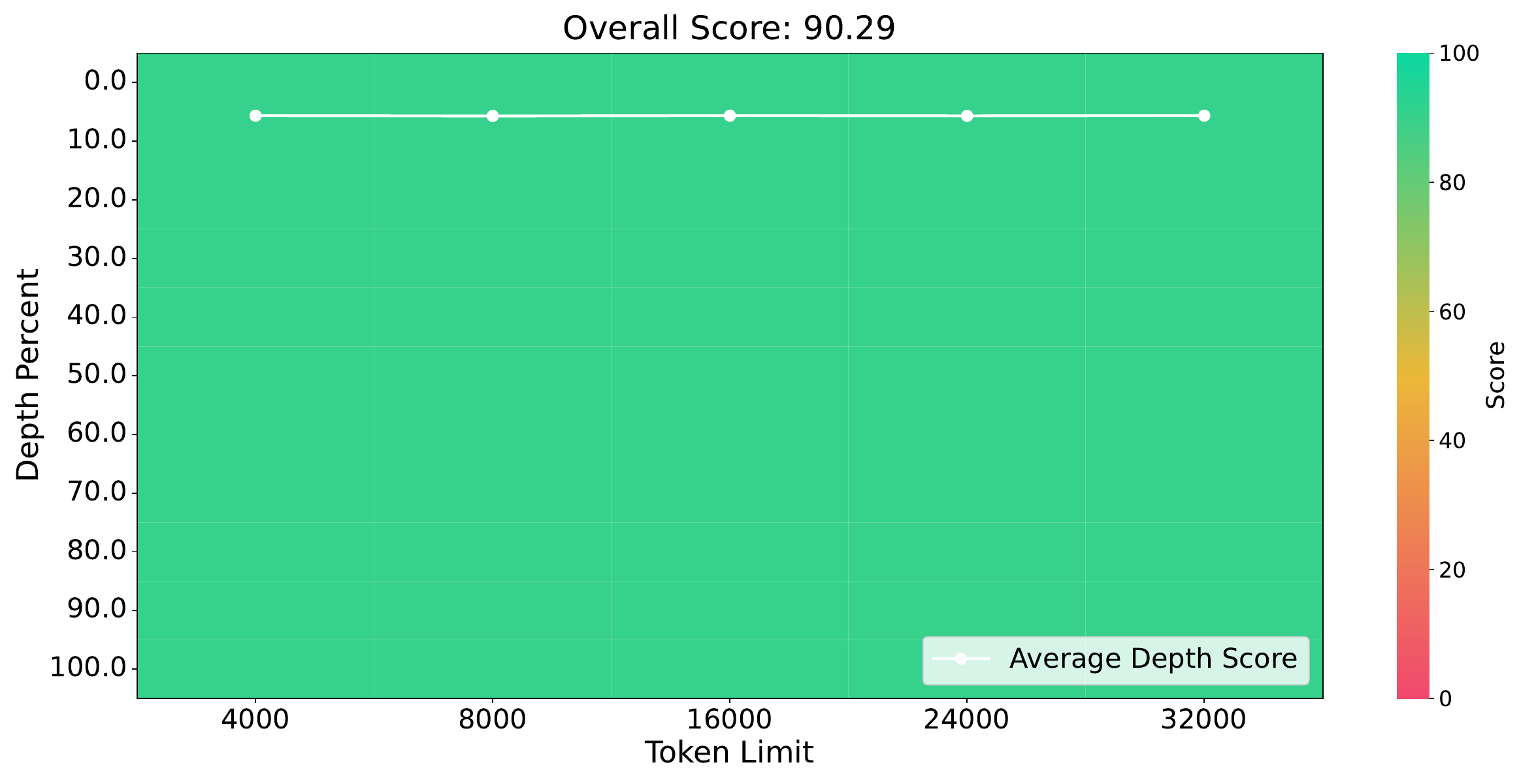}
        \caption{SnapKV}
        \label{niah_llama3_2_3b_snapkv}
    \end{subfigure}
    \vskip\baselineskip
    \begin{subfigure}[b]{0.32\linewidth}
        \centering
        \includegraphics[width=\linewidth]{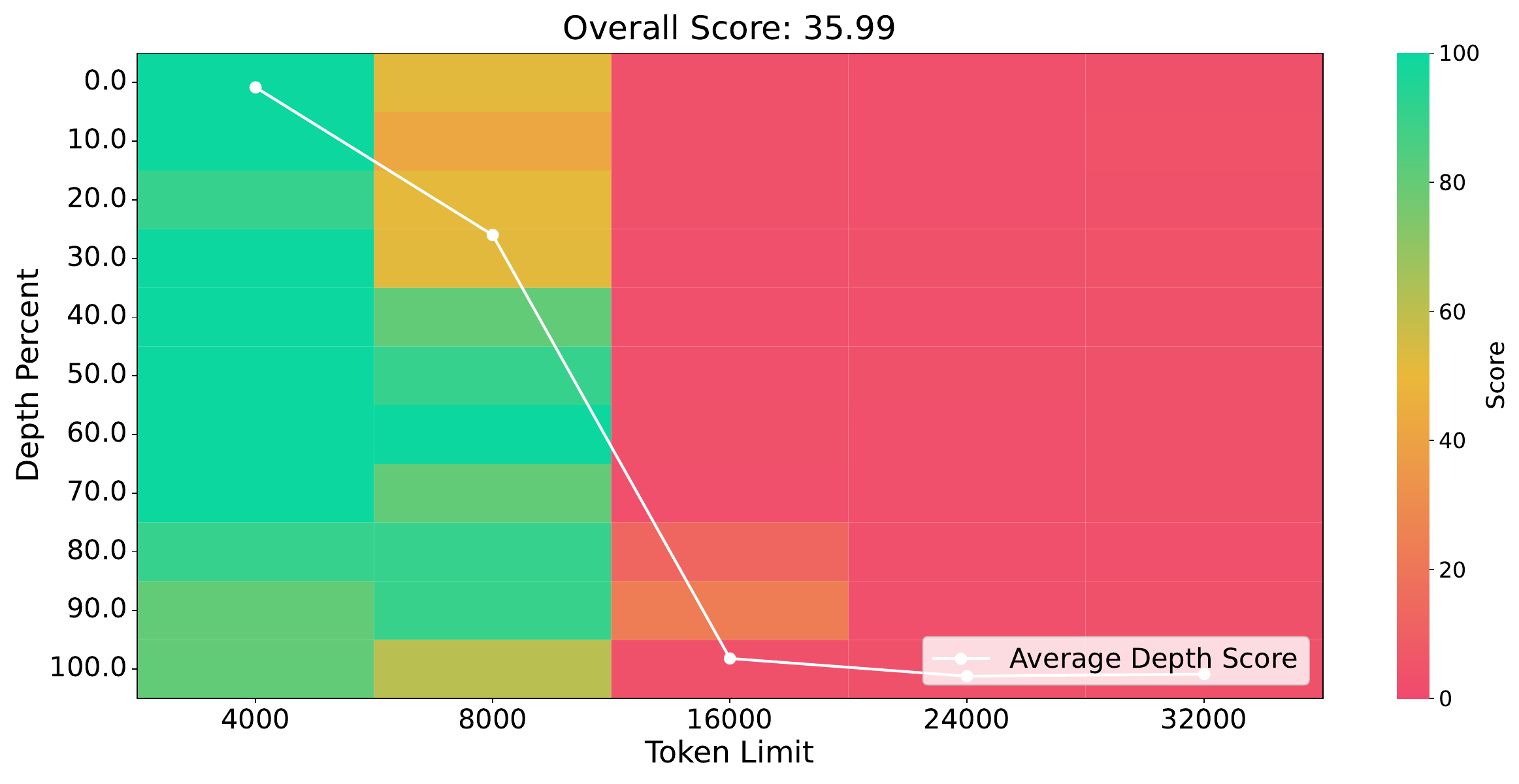}
        \caption{Palu}
        \label{niah_llama3_2_3b_palu}
    \end{subfigure}
    \hfill
    \begin{subfigure}[b]{0.32\linewidth}
        \centering
        \includegraphics[width=\linewidth]{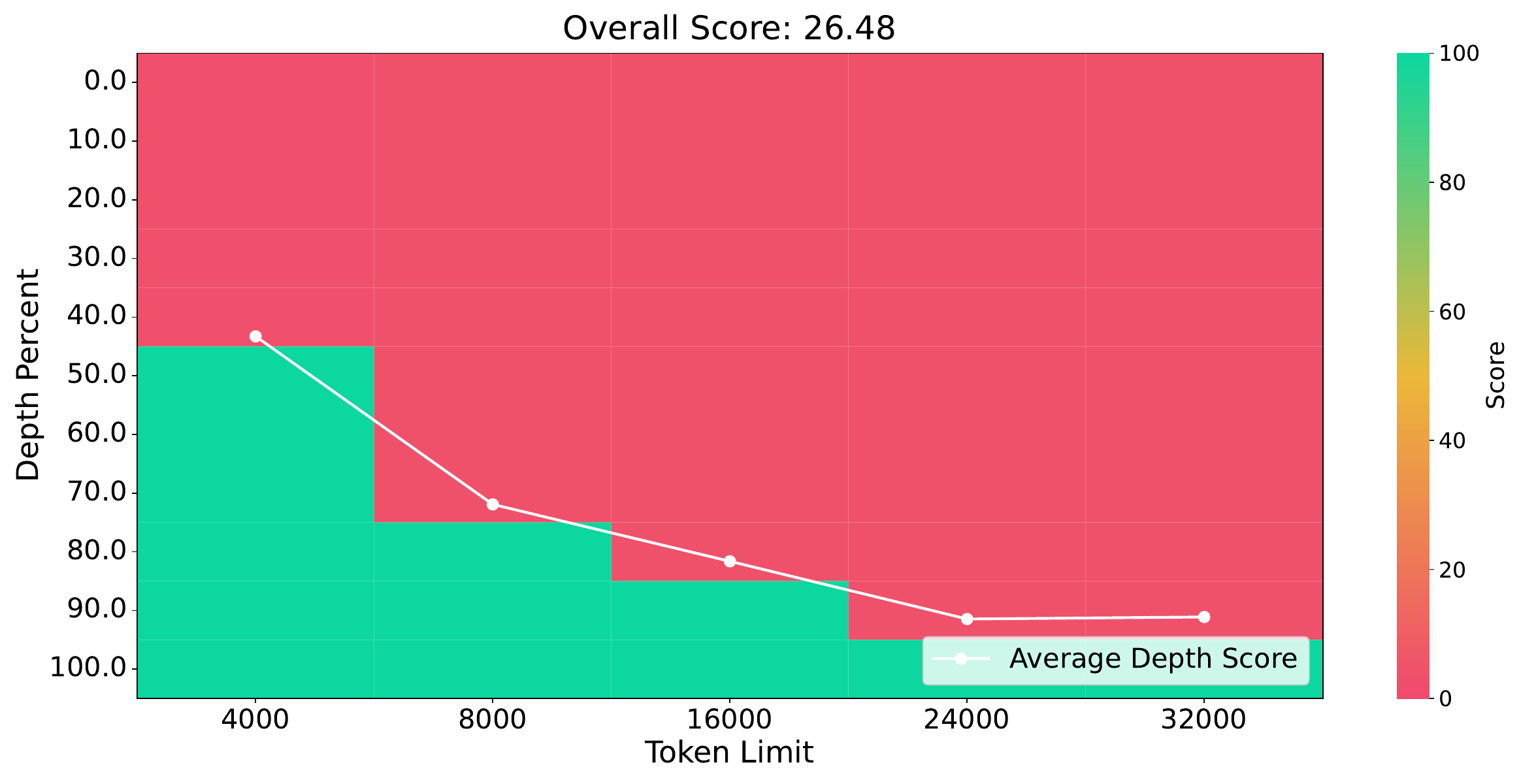}
        \caption{KIVI}
        \label{niah_llama3_2_3b_kivi}
    \end{subfigure}
    \hfill
    \begin{subfigure}[b]{0.32\linewidth}
        \centering
        \includegraphics[width=\linewidth]{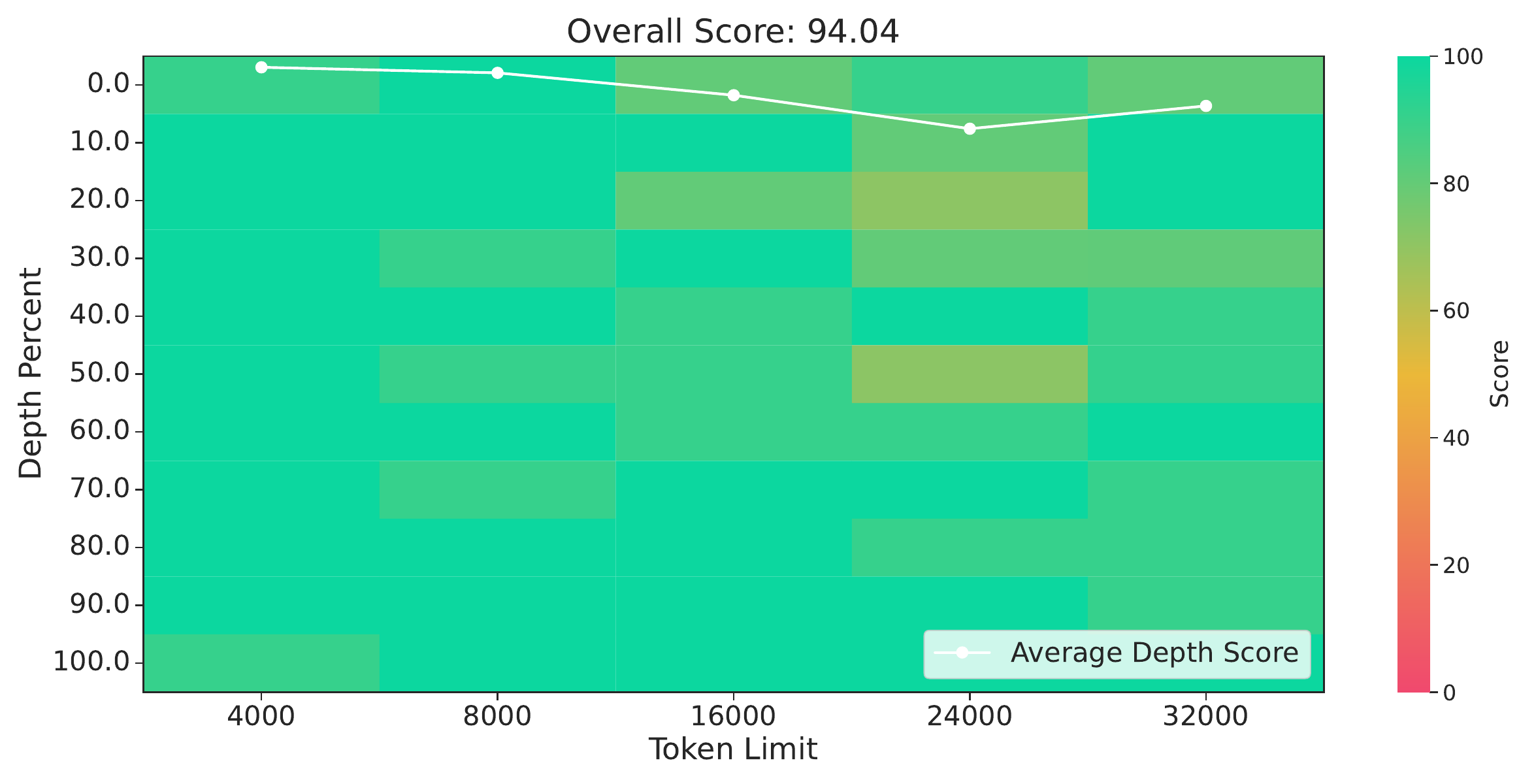}
        \caption{{\method} (ours)}
        \label{niah_llama3_2_3b_ours}
    \end{subfigure}
    \caption{Results of LLaMA3.2-3B~\citep{meta2024introducing} on Needle-In-A-Haystack~\citep{niah}. {\method} achieves a highest average score over StreamingLLM~\citep{xiaoefficient}, SnapKV~\citep{li2024snapkv}, Palu~\citep{chang2024palu}, and KIVI~\citep{liu2024kivi} and shows closest performance with LLMs with full attention. \label{niah_llama3_2_3b}}
\end{figure*}

For LongBench as shown in Tables~\ref{longbench1} and \ref{longbench2}, our {\method} achieves performance closest to the original model on LLaMA3.2-3B and is slightly inferior to SnapKV on LLaMA3.1-8B. For the NIAH task as shown in Figure~\ref{niah_llama3_1_8b} and \ref{niah_llama3_2_3b}, we similarly achieve performance closest to the original pretrained models at 32k context length. While SnapKV is theoretically suitable for retrieval tasks like NIAH, it still exhibits recall errors. Though Palu and KIVI maintain stable attention approximation under moderate compression, 30-50\%, they show significant performance degradation at ~75\% compression due to insufficient granular analysis of KV cache features. In contrast, our {\method} optimizes compression by identifying and preserving KV dimensions insensitive to compression, thereby maximally retaining the long-context capabilities and demonstrating superiority across both models and benchmarks.


In addition to comparisons in downstream performance, we will also conduct efficiency experiments. Since our custom kernel {\kernel} is still in progress, we will report this in detail in the next version of the paper.





\section{Discussion}\label{sec:diss}

\subsection{Choice of Basis Functions}\label{sec:basis}

\begin{table}[tb]
\tabcolsep=0.1cm
\centering\small
\begin{tabular}{lccccc}
\toprule
& \textbf{MK1} & \textbf{MK2} & \textbf{MK3} & \textbf{MV} & \textbf{Avg.} \\
\midrule
\textcolor{gray}{\textbf{\textit{LLaMA3.2-3B}}} & \textcolor{gray}{99.00} & \textcolor{gray}{100.00} & \textcolor{gray}{99.00} & \textcolor{gray}{100.00} & \textcolor{gray}{99.50} \\
+ FourierT & 99.00 & 99.00 & \textbf{99.00} & \textbf{96.00} & \textbf{98.25} \\
+ LegT & 89.00 & 93.00 & 50.00 & 93.75 & 81.44 \\
\midrule
+ uniform & \textbf{100.00} & 99.00 & 98.00 & 90.50 & 96.88 \\
+ KV inv. & 98.00 & \textbf{100.00} & 98.00 & 93.00 & 97.25 \\
+ layer inv. & 99.00 & 98.00 & 93.00 & 86.00 & 94.00 \\
\bottomrule
\end{tabular}
\caption{Validation of basis function and compression schema in LLaMA3.2-3B~\citep{meta2024introducing}\label{ruler}}
\end{table}

\begin{figure*}[!tb]
    \begin{subfigure}[b]{0.48\linewidth}
        \centering
        \includegraphics[width=\linewidth]{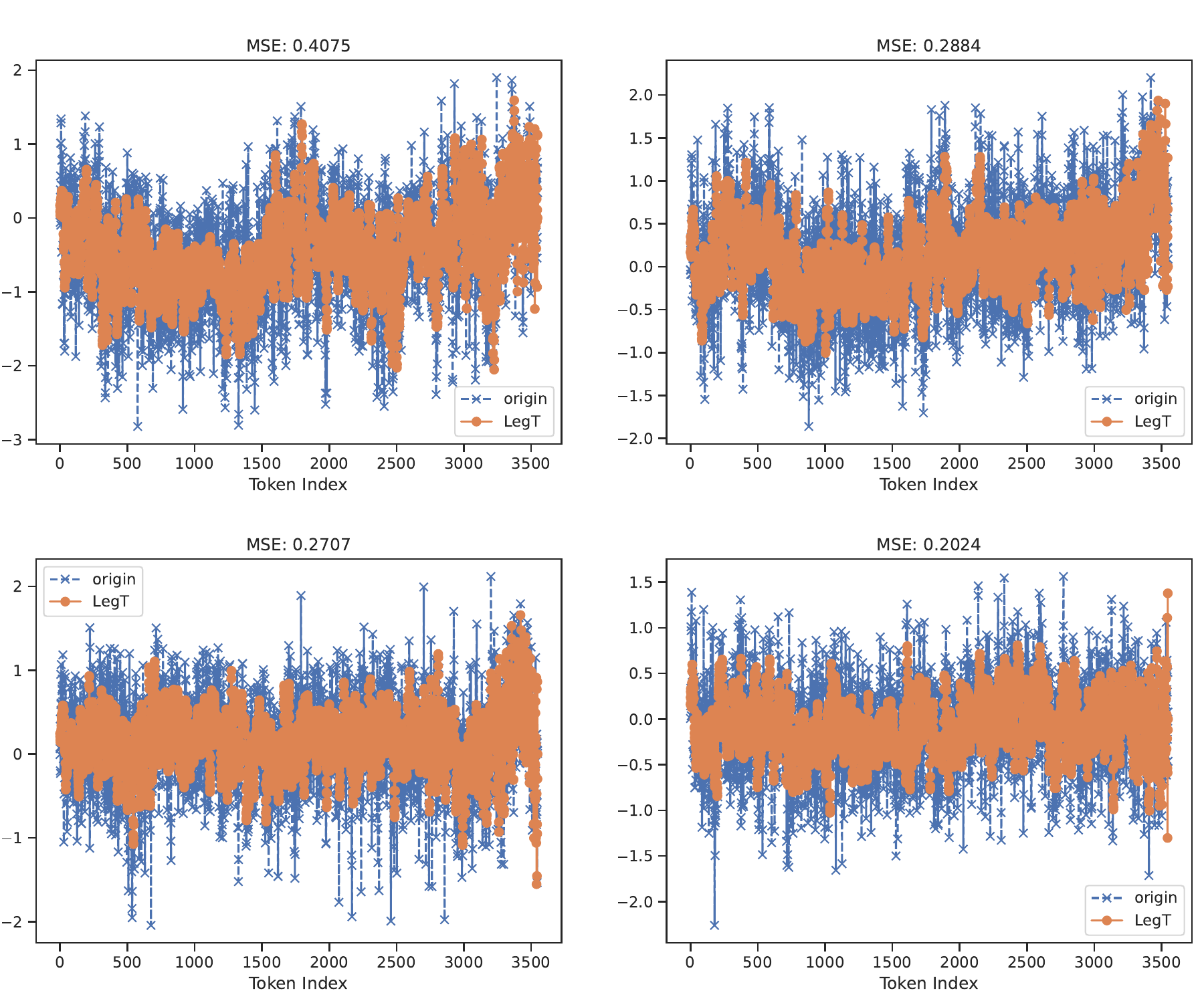}
        \caption{Cache reconstruction by HiPPO-LegT}
        \label{hippo_legt}
    \end{subfigure}
    \hfill
    \begin{subfigure}[b]{0.48\linewidth}
        \centering
        \includegraphics[width=\linewidth]{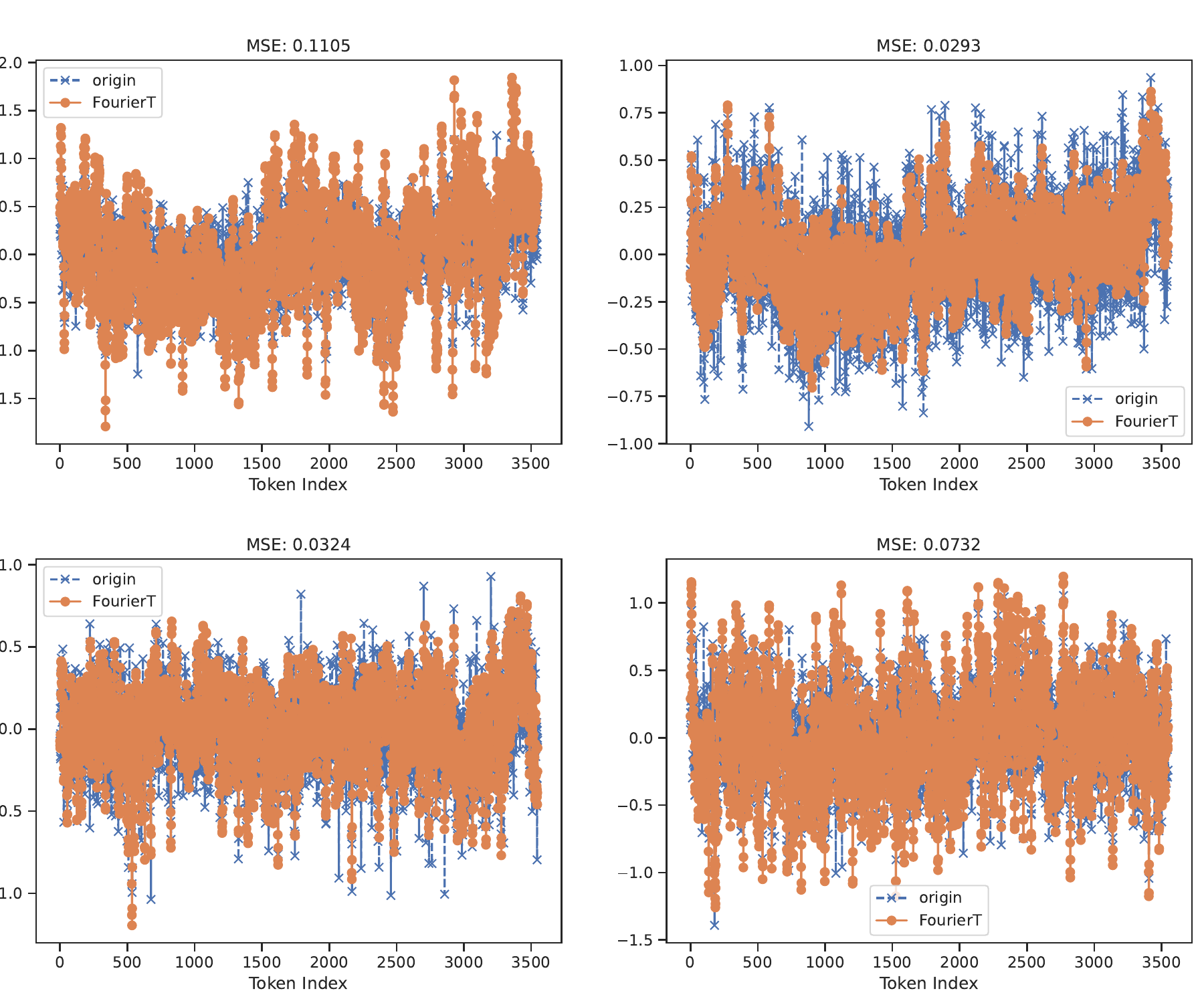}
        \caption{Cache reconstruction by HiPPO-FourierT}
        \label{hippo_fouriert}
    \end{subfigure}
    \caption{Visualization of KV cache reconstruction in LLaMA3.2-3B~\citep{meta2024introducing} for different basis functions, LegT and FourierT under HiPPO framework~\citep{gu2020hippo}. FourierT outperforms LegT in cache reconstruction.\label{fig:basis_cmp}}
\end{figure*}

Although {\method} employs HiPPO-FourierT for compression, \citet{gu2020hippo} proposes and claims polynomial basis functions like LegT with superior performance. While maintaining identical sliding window sizes, we compare LegT and FourierT in reconstructing KV caches from LLMs. As illustrated in Figure~\ref{fig:basis_cmp}, we evaluate their reconstruction effects on 4 randomly selected KV cache dimensions from layer 0 of LLaMA3.2. Under equivalent state dimensions\footnote{FourierT uses lower-order basis functions since {\method}'s state size is twice the number of states}, FourierT consistently achieves lower reconstruction loss than LegT.

We further evaluate FourierT and LegT compression on LLaMA3.2-3B using more discriminative NIAH variants, Multi-Key NIAH (MK) and Multi-Value NIAH (MV)~\citep{hsieh2024ruler} in 4k context length. For fair comparison, we employ the same method to identify dimensions suitable for LegT compression and apply an identical compression schema. Results in Table~\ref{ruler} show FourierT still performs better, demonstrating that FourierT offers better parallelizability for compression efficiency and performance in downstream evaluation.

\begin{figure*}[!th]
    \begin{subfigure}[b]{0.24\linewidth}
        \centering
        \includegraphics[width=\linewidth]{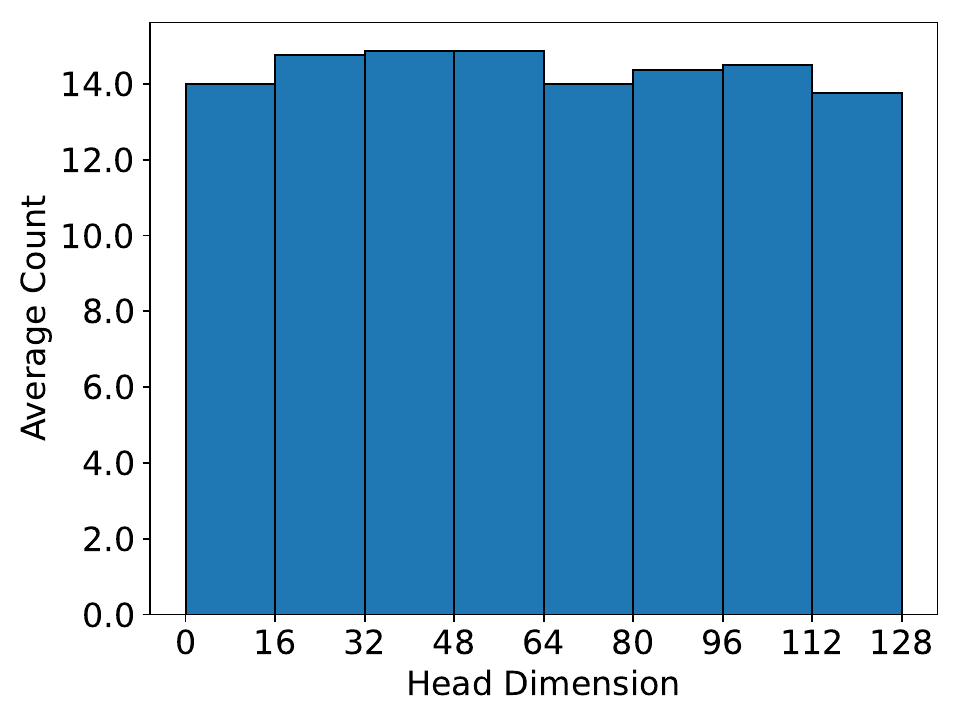}
        \caption{Layer 0 in LLaMA3.1-8B}
        \label{llama_3_1_8b_layer0_dim}
    \end{subfigure}
    \hfill
    \begin{subfigure}[b]{0.24\linewidth}
        \centering
        \includegraphics[width=\linewidth]{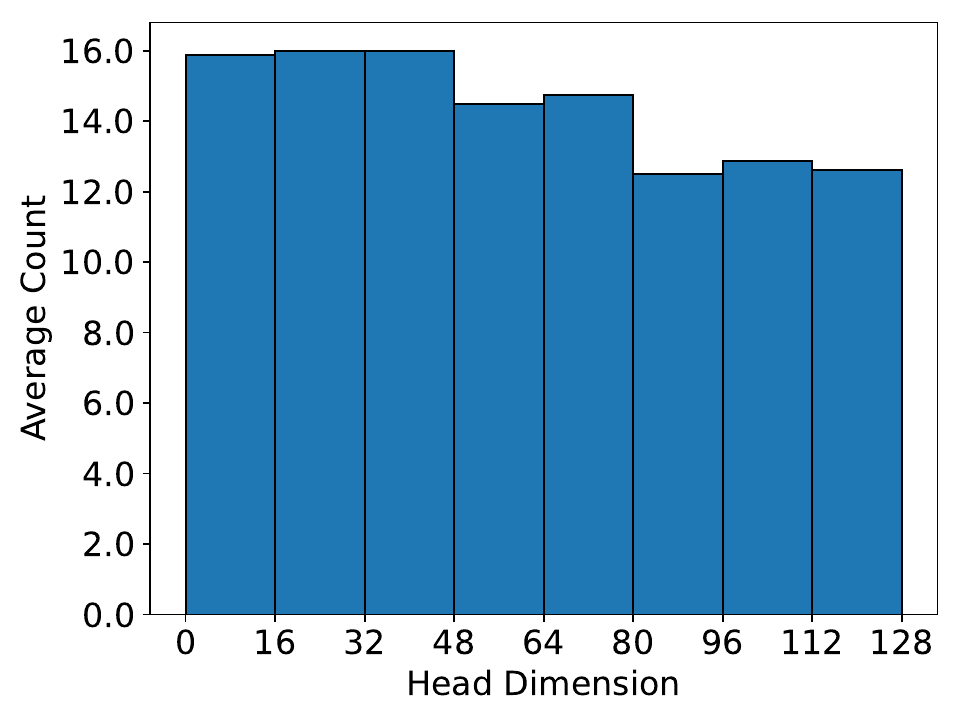}
        \caption{Layer 2 in LLaMA3.1-8B}
        \label{llama_3_1_8b_layer2_dim}
    \end{subfigure}
    \hfill
    \begin{subfigure}[b]{0.24\linewidth}
        \centering
        \includegraphics[width=\linewidth]{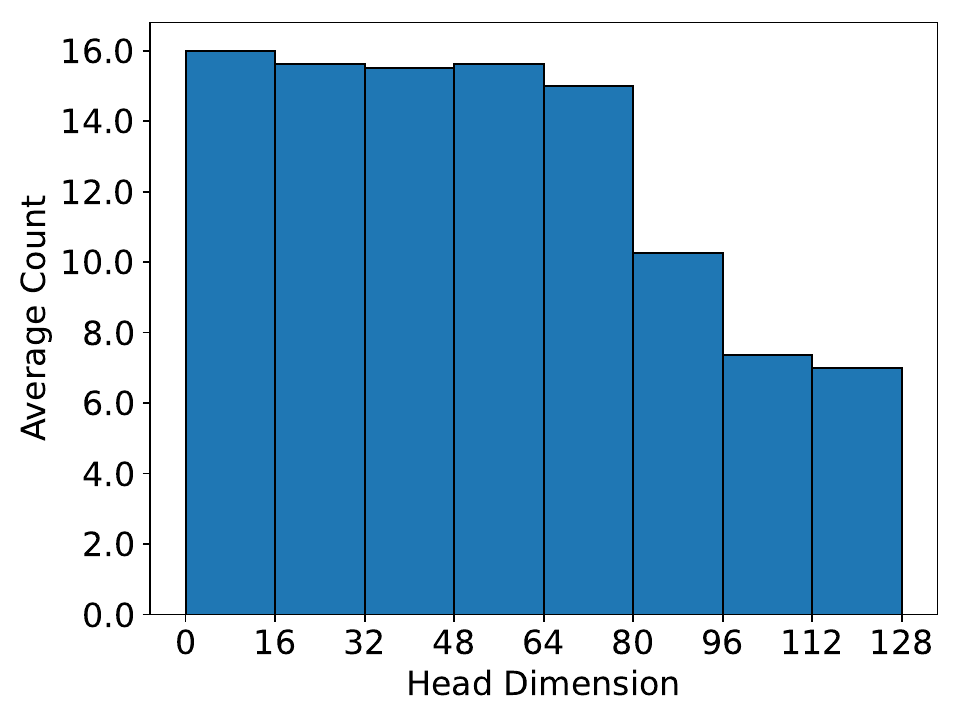}
        \caption{Layer 15 in LLaMA3.1-8B}
        \label{llama_3_1_8b_layer15_dim}
    \end{subfigure}
    \hfill
    \begin{subfigure}[b]{0.24\linewidth}
        \centering
        \includegraphics[width=\linewidth]{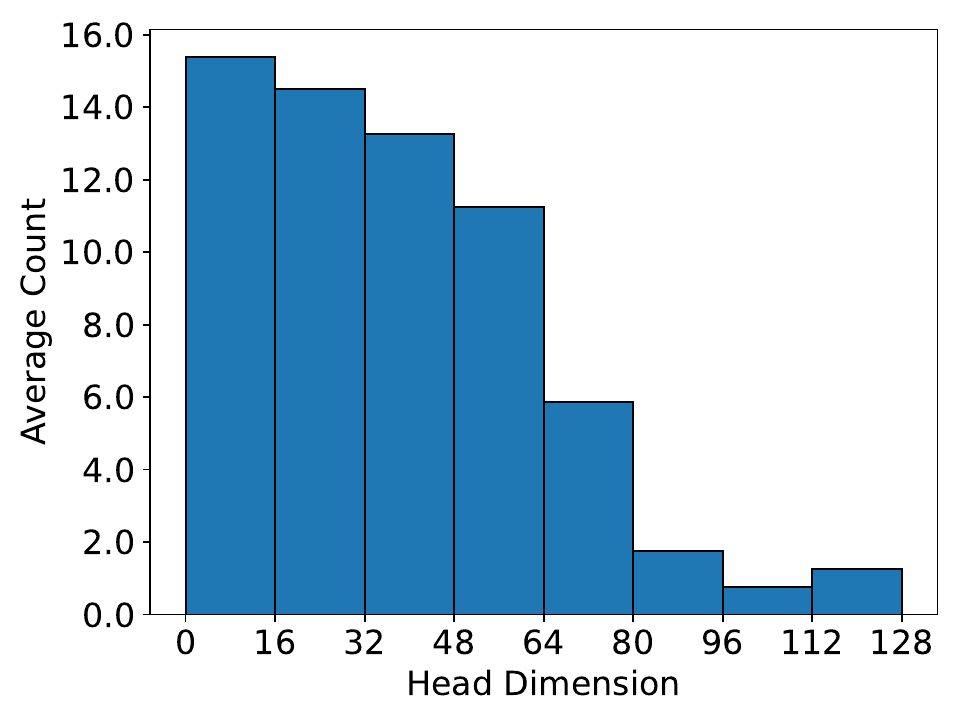}
        \caption{Layer 30 in LLaMA3.1-8B}
        \label{llama_3_1_8b_layer30_dim}
    \end{subfigure}
    \begin{subfigure}[b]{0.24\linewidth}
        \centering
        \includegraphics[width=\linewidth]{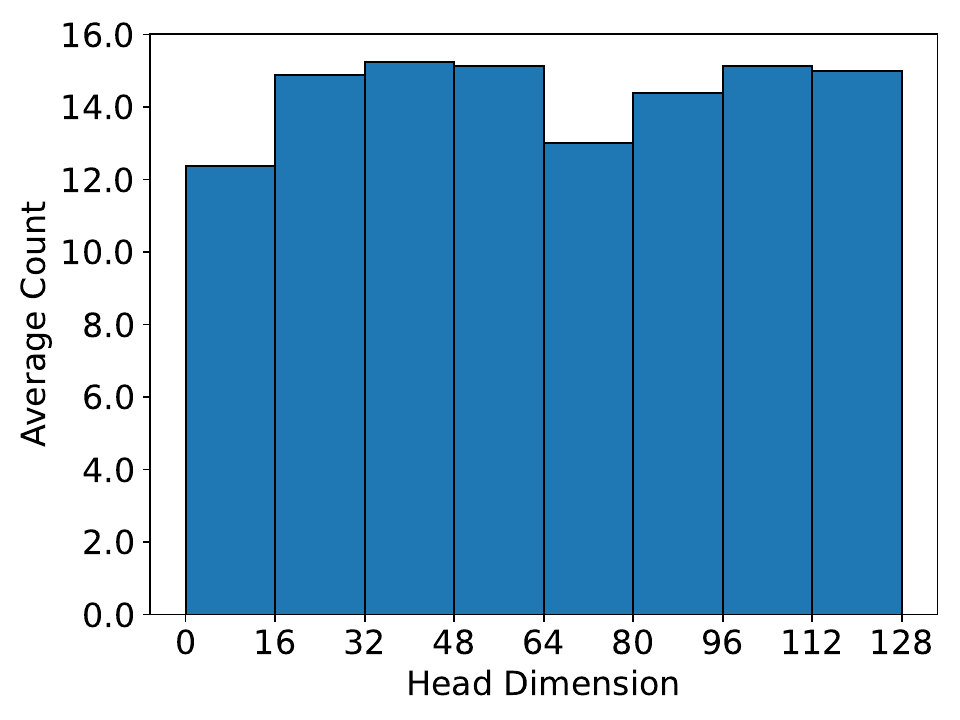}
        \caption{Layer 0 in LLaMA3.2-3B}
        \label{llama_3_2_3b_layer0_dim}
    \end{subfigure}
    \hfill
    \begin{subfigure}[b]{0.24\linewidth}
        \centering
        \includegraphics[width=\linewidth]{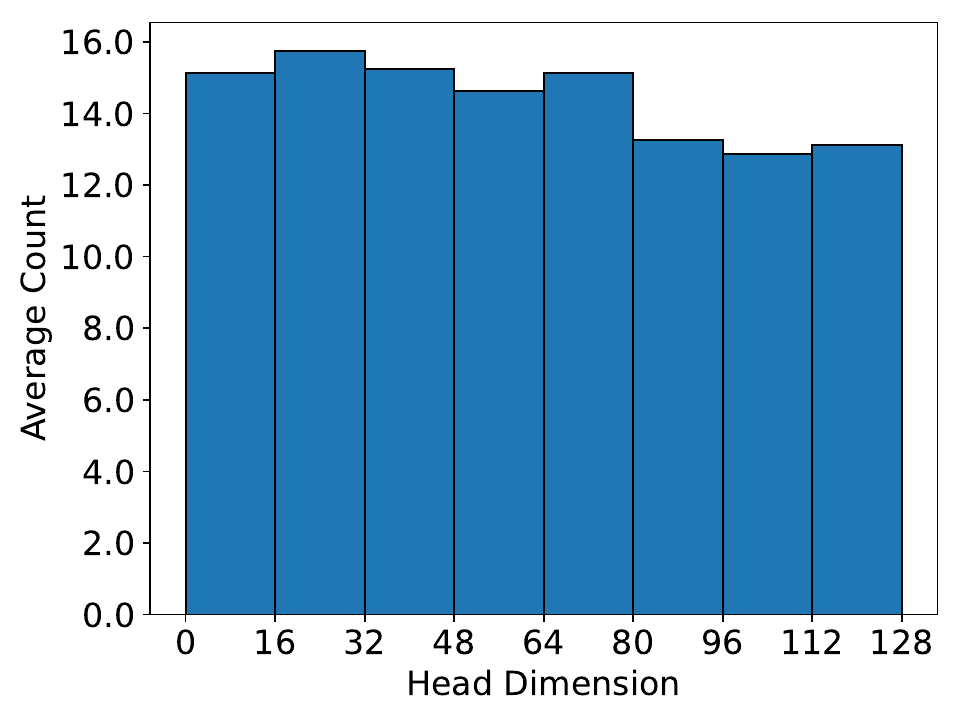}
        \caption{Layer 2 in LLaMA3.2-3B}
        \label{llama_3_2_3b_layer2_dim}
    \end{subfigure}
    \hfill
    \begin{subfigure}[b]{0.24\linewidth}
        \centering
        \includegraphics[width=\linewidth]{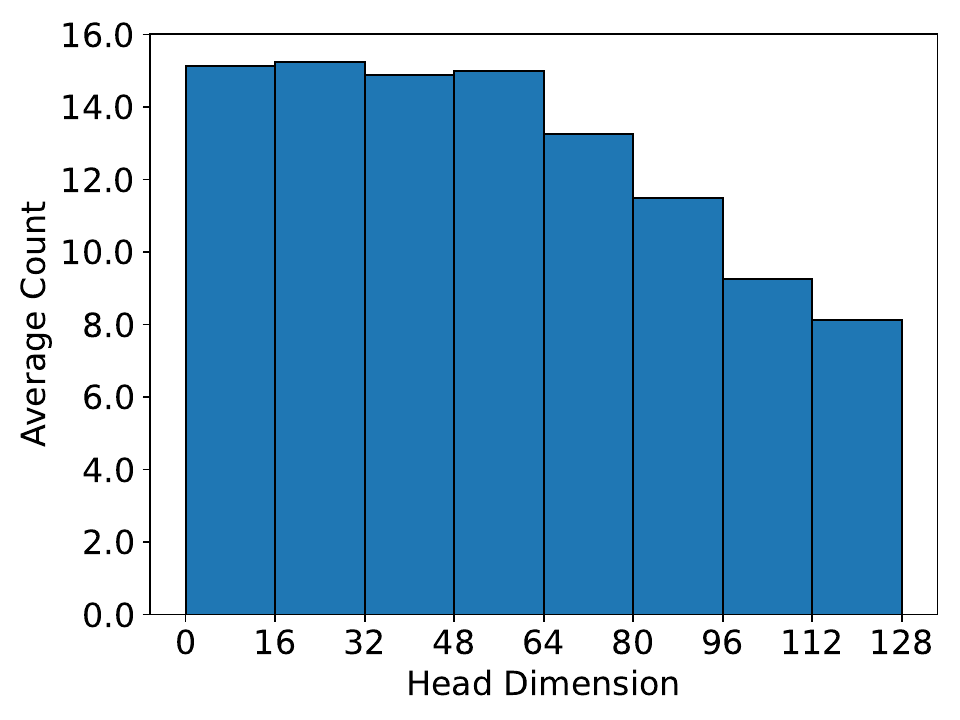}
        \caption{Layer 13 in LLaMA3.2-3B}
        \label{llama_3_2_3b_layer13_dim}
    \end{subfigure}
    \hfill
    \begin{subfigure}[b]{0.24\linewidth}
        \centering
        \includegraphics[width=\linewidth]{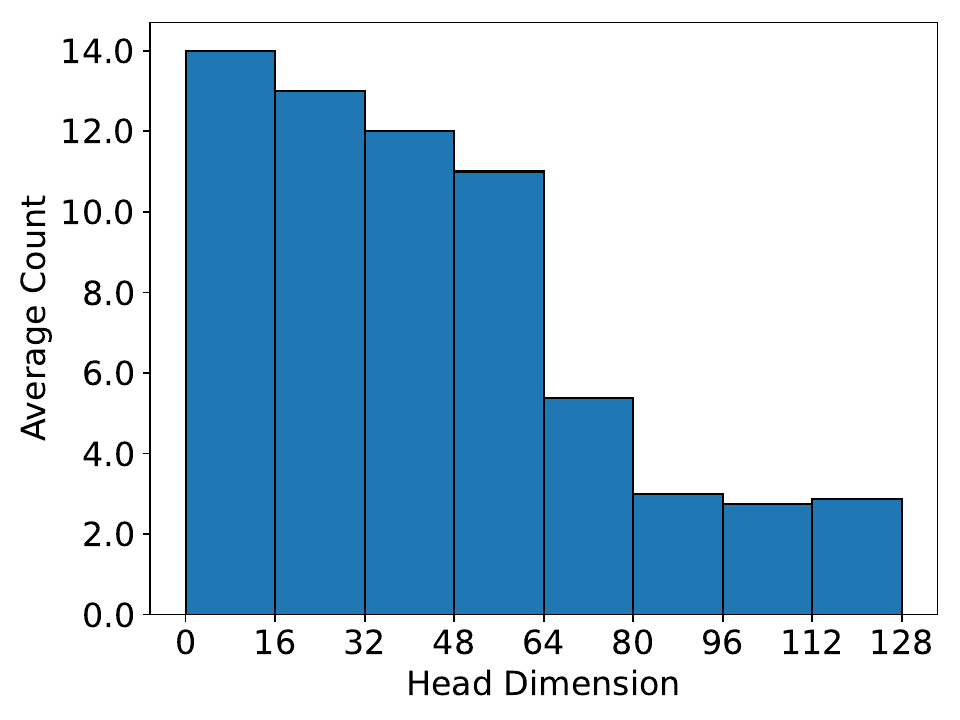}
        \caption{Layer 26 in LLaMA3.2-3B}
        \label{llama_3_2_3b_layer26_dim}
    \end{subfigure}
    \caption{The statistics of each dimension selected for compression, averaged across attention heads in different layers, grouped every 16 dimensions, in LLaMA3.1-8B~\citep{dubey2024llama} and LLaMA3.2-3B~\citep{meta2024introducing}.\label{fig:dim_dist}}
\end{figure*}

\subsection{Ablation on Compression Schema}\label{sec:ratio}

As mentioned in Section~\ref{sec:select}, we propose a more fine-grained compression scheme based on additional observations of the KV cache. As shown in the Table~\ref{ruler}, we compare three approaches: uniform compression across all layers and between KV (uniform), inverted KV compression schema by K-priority over V (KV inv.), and inverted layer-wise compression schema by upper-layer priority over lower-layer (layer inv.). Results demonstrate that our original V-priority and lower-layer-priority compression schema achieves superior performance on discriminative NIAH variants. This further illustrates that frequency-based sequence-wise KV cache compression exhibits fundamentally different optimization characteristics compared to conventional KV token eviction methods~\citep{cai2024pyramidkv,xing2024pyramiddrop}.

\subsection{Compressed Dimension Distribution}\label{sec:dim}

Finally, we analyze the compressed dimensions selected by our {\method}. We count the number of each dimension selected for compression, averaged across attention heads in different layers, grouped every 16 dimensions. Results in Figure~\ref{fig:dim_dist} show that in both LLaMA3.1-8B and LLaMA3.2-3B, starting from layer 2, lower dimensions are more frequently compressed while the upper dimensions tend to preserve complete temporal information in our {\method}. This phenomenon is more evident in upper layers, where fewer dimensions are chosen to be compressed. As illustrated in Figure~\ref{fig:heatmap_niah}, these uncompressed upper dimensions primarily contribute to forming attention sinks and capturing long-context semantic relationships, thus requiring complete retention, whereas other dimensions can be stored with limited length.

\section{Conclusion}

We propose {\method}, a novel KV cache optimization approach that compresses long-context-insensitive dimensions without sacrificing contextual awareness based on an interesting phenomenon in transformer head dimensions, that lower dimensions capture local features, while upper ones capture long-context dependencies. Inspired by HiPPO, we optimize the long-context-insensitive KV cache through a translated Fourier transform into fixed-length states in the prefilling phase and reconstruct the KV cache in the decoding phase. {\method} shows the best performance on the LLaMA Series in LongBench and NIAH on average. We are trying to improve the efficiency of {\method} through a customized Triton-based kernel, {\kernel}, eliminating intermediate read-write operations and effectively reducing memory overhead.  

\section*{Limitations}


We will continue optimizing our customized Triton kernel \textbf{\kernel}, which injects KV cache decomposition in FlashAttention and FlashDecoding~\citep{daoflashattention}, minimizing memory overhead via streamlined read-write operations. Moreover, our performance still shows gaps compared to the pre-trained model. These aspects will be thoroughly investigated in future work.

\section*{Acknowledgments}

We thank Jian Yuan from Shanghai Jiao Tong University for assisting with experimental verification.


\bibliography{custom}

\end{document}